\documentclass[10pt,twocolumn,letterpaper]{article}

\usepackage[review]{cvpr}
\usepackage{times}
\usepackage{epsfig}
\usepackage{graphicx}
\usepackage{amsmath}
\usepackage{amssymb}
\usepackage{multirow}
\usepackage{cite}
\usepackage{makecell}
\usepackage{tabularx}
\usepackage{lscape}
\usepackage{pdflscape}
\usepackage{subcaption}
\usepackage{booktabs}
\usepackage{enumitem}
\usepackage[
  separate-uncertainty = true,
  multi-part-units = repeat
]{siunitx}
\newcolumntype{P}[1]{>{\centering\arraybackslash}p{#1}}

\usepackage[pagebackref=true,breaklinks=true,letterpaper=true,colorlinks,bookmarks=false]{hyperref}

\begin{document}
\title{Deepfake Synthesis vs. Detection: An Uneven Contest}

\author{
Md. Tarek Hasan\textsuperscript{1},
Sanjay Saha\textsuperscript{2},
Shaojing Fan\textsuperscript{2},
Swakkhar Shatabda\textsuperscript{3},
Terence Sim\textsuperscript{2} \\
\\
\textsuperscript{1}United International University, Bangladesh \\
{\tt\small tarek@cse.uiu.ac.bd} \\
\\
\textsuperscript{2}National University of Singapore, Singapore \\
{\tt\small sanjaysaha@u.nus.edu,
fanshaojing@nus.edu.sg,
terence.sim@nus.edu.sg} \\
\\
\textsuperscript{3}BRAC University, Bangladesh \\
{\tt\small swakkhar.shatabda@bracu.ac.bd}
}

\maketitle


\begin{abstract}
The rapid advancement of deepfake technology has significantly elevated the realism and accessibility of synthetic media. Emerging techniques, such as diffusion-based models and Neural Radiance Fields (NeRF), alongside enhancements in traditional Generative Adversarial Networks (GANs), have contributed to the sophisticated generation of deepfake videos. Concurrently, deepfake detection methods have seen notable progress, driven by innovations in Transformer architectures, contrastive learning, and other machine learning approaches. In this study, we conduct a 
comprehensive empirical analysis of state-of-the-art deepfake detection techniques, including human evaluation experiments against cutting-edge synthesis methods.
Our findings highlight a concerning trend: many state-of-the-art detection models exhibit markedly poor performance when challenged with deepfakes produced by modern synthesis techniques, including poor performance by human participants against the best quality deepfakes. Through extensive experimentation, we provide evidence that underscores the urgent need for continued refinement of detection models to keep pace with the evolving capabilities of deepfake generation technologies. This research emphasizes the critical gap between current detection methodologies and the sophistication of new generation techniques, calling for intensified efforts in this crucial area of study.
\end{abstract}


%
\section{Introduction}

The proliferation of deepfake technology has fundamentally altered the landscape of digital media, enabling the creation of highly realistic synthetic videos that are increasingly difficult to distinguish from authentic content. Originally popularized through the use of Generative Adversarial Networks (GANs) \cite{goodfellow2014generative}, deepfake generation techniques have evolved rapidly, incorporating advanced methodologies such as diffusion-based models \cite{sohl2015deep, ho2020denoising} and Neural Radiance Fields (NeRF) \cite{mildenhall2021nerf}. These innovations have significantly enhanced the quality and complexity of deepfakes, making them more convincing and harder to detect.

While the progress in deepfake synthesis is remarkable, it also poses substantial challenges for detection models, which are critical for mitigating the potential misuse of this technology. Recent advances in machine learning, particularly in Transformer models \cite{vaswani2017attention, dosovitskiy2020image} and contrastive learning techniques \cite{chen2020simple, he2020momentum}, have led to the development of more sophisticated deepfake detection methods. However, despite these improvements, there is a growing concern that detection models may not be advancing quickly enough to counteract the rapid evolution of deepfake generation methods.

Talking-head synthesis and face reenactment have evolved rapidly through advancements in generative modeling, particularly with diffusion-based and neural rendering techniques. Early works like Wang et al.~\cite{wang2021one} and Zhao et al.~\cite{zhao2022thin} introduced efficient video generation and motion transfer methods. Diffusion-based systems such as DiffusedHeads~\cite{stypulkowski2024diffused}, DreamTalk~\cite{ma2023dreamtalk}, and AniFaceDiff~\cite{chen2024anifacediff} improved expressiveness and identity fidelity using noise modeling and conditioning strategies. VASA~\cite{xu2024vasa} further advanced the field with affective dynamics, while SyncTalk~\cite{peng2024synctalk} integrated NeRFs for lip and head synchronization. HyperReenact~\cite{bounareli2023hyperreenact} leveraged StyleGAN2 with hypernetworks to support extreme pose reenactment and identity preservation. Together, these models mark a shift from static keypoint-based methods toward highly expressive, real-time, and emotionally aware talking-face generation.

Face tampering detection has progressed from shallow CNNs to advanced architectures incorporating attention, disentanglement, and frequency-domain features. Foundational models like MesoNet~\cite{afchar2018mesonet} and XceptionNet~\cite{rossler2019faceforensics++} emphasized mesoscopic cues and separable convolutions. Subsequent works explored capsule networks~\cite{nguyen2019capsule}, EfficientNet scaling~\cite{tan2019efficientnet}, and attention-guided detection~\cite{dang2020detection} to enhance feature discrimination. Recent models such as CoRe~\cite{ni2022core}, UCF~\cite{yan2023ucf}, and methods by Luo et al.~\cite{luo2021generalizing} and Cao et al.~\cite{cao2022end} tackle cross-domain generalization through representation disentanglement, noise-based cues, and graph reasoning. These approaches address overfitting and unseen forgery styles, achieving improved robustness in detecting increasingly sophisticated manipulations.

This study empirically evaluates state-of-the-art deepfake detection methods against cutting-edge generation techniques. Through comprehensive experiments, we reveal that many current models struggle with modern deepfakes, exposing critical limitations and highlighting areas for improvement.

These findings carry significant implications for digital content integrity in domains such as media, security, and privacy. Prior work~\cite{nguyen2019multi, rossler2019faceforensics++} has noted the growing difficulty of detecting increasingly realistic fakes. Our results emphasize the urgent need for continuous innovation to ensure detection systems can adapt to evolving threats. Recent surveys often lack empirical validation and real-world relevance. To address this, we conduct extensive evaluations of the latest detection and synthesis methods, supplemented by a human study assessing the effectiveness of human judgment in identifying deepfakes.

Our findings reveal surprising insights into the current state of deepfake detection, highlighting significant shortcomings in the existing literature that need immediate attention from researchers. The results underscore the urgency of developing more robust detection methods to keep pace with the rapidly evolving synthesis technologies.

Our key contributions are as follows:

\begin{itemize}[leftmargin=0pt,itemsep=2pt, topsep=2pt]
    \item Comprehensive Empirical Evaluation: We provide a thorough empirical analysis of the latest state-of-the-art deepfake generation and detection methods, offering valuable insights into their performance and limitations.
    \item Human Evaluation Study: We conduct a detailed human survey to assess the detection capabilities of individuals when faced with highly realistic deepfakes, providing critical data on human performance in this domain.
    \item Future Research Directions: We offer clear guidance for future research, identifying specific areas where improvements are needed to bridge the gap between current detection methods and the advanced synthesis techniques that challenge them.
\end{itemize}
This study not only advances our understanding of the effectiveness of deepfake detection technologies but also sets the stage for future research aimed at enhancing both algorithmic and human detection capabilities in the face of increasingly sophisticated deepfake content.

\section{Method}

\begin{figure*}[h]
\centering
\includegraphics[width=\textwidth]{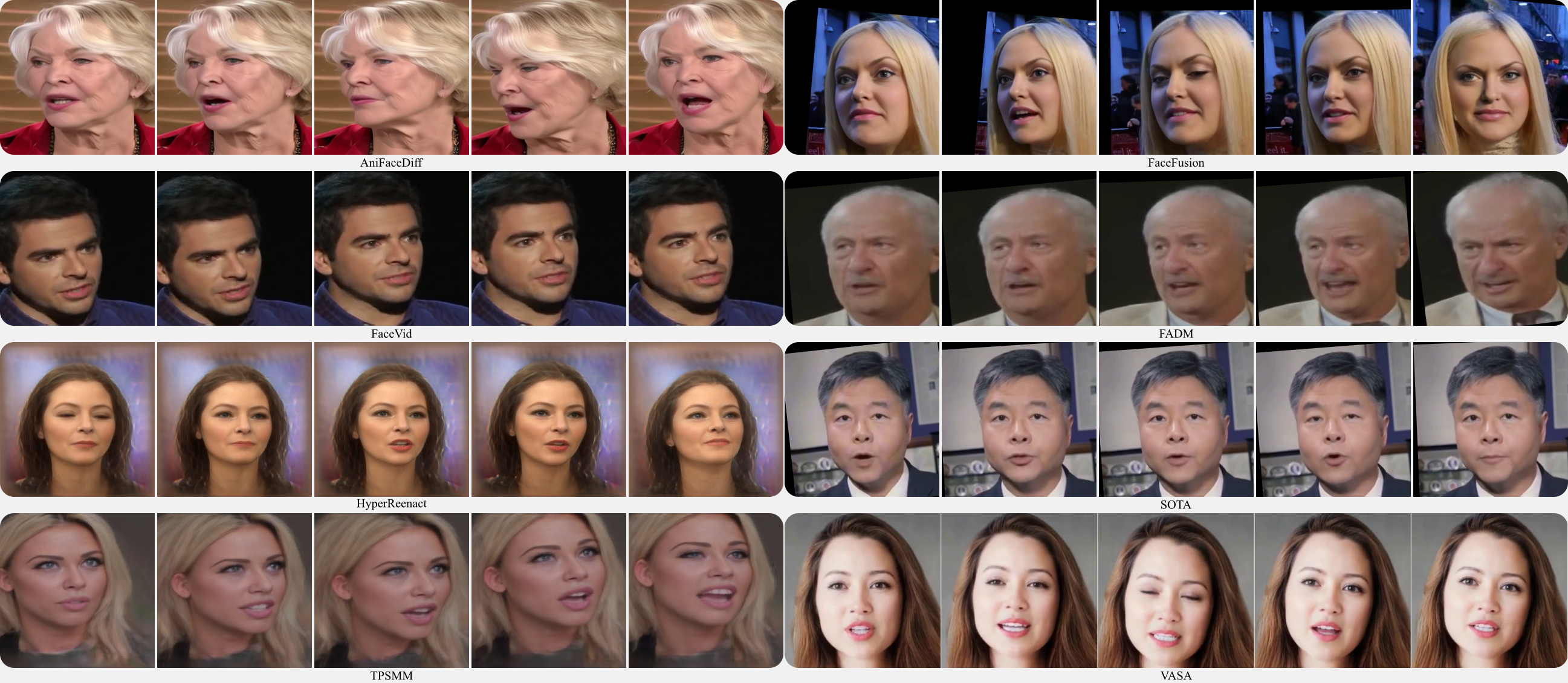}
\caption{Example sequence of frames from videos generated using the synthesis methods used in the paper. This figure show the high quality and realistic deepfake videos used in this study. Method in the figure from top to bottom respectively: AniFaceDiff \cite{chen2024anifacediff}, FaceFusion \cite{facefusion2024}, FaceVid \cite{wang2021one}, FADM \cite{zeng2023face}, HyperReenact \cite{bounareli2023hyperreenact}, Synctalk \cite{peng2024synctalk}, TPSMM \cite{zhao2022thin}, and VASA-1 \cite{xu2024vasa}.}
\label{fig:synthesis-methods}
\end{figure*}

In this study, we adopt a systematic empirical approach to evaluate both deepfake generation and detection methods. To ensure consistency, the same set of silent deepfake videos—without audio to focus attention on visual cues—was used for both algorithmic and human evaluations. Our methodology comprises three key stages. First, we examine the deepfake synthesis techniques, including diffusion models, GAN variants, and NeRF-based approaches, analyzing their strengths, limitations, and influence on realism and detectability. Next, we assess state-of-the-art detection models, such as those based on Transformer architectures and contrastive learning, detailing their design and performance. Finally, we conduct a human evaluation using the same video stimuli to compare human and algorithmic performance, offering insights into the perceptual cues humans use and the challenges of real-world deepfake detection.

\begin{table*}[]
\scriptsize
\centering
\footnotesize
\def\arraystretch{1.2}
\caption{Comprehensive summary of deepfake generation techniques, including methodology, dataset scale, data modality, and technical contributions.}
\begin{tabular}{P{1.6cm}P{1.4cm}P{2cm}P{2.8cm}P{1.4cm}P{2cm}p{4cm}}
\hline
\textbf{Name} & \textbf{Publishing} & \textbf{Method} & \textbf{Dataset Used} & \textbf{Type} & \textbf{Number of Images/Videos} & \textbf{Highlights} \\ \hline

FaceVid \cite{wang2021one} & CVPR'21 & GAN & VoxCeleb2, TalkingHead-1KH & Image-to-video & 1M (VoxCeleb2), 180K (TalkingHead-1KH) & Generates realistic talking-heads from a single image, separates appearance and motion using keypoints, achieves high fidelity using warping-based generation \\

TPSMM \cite{zhao2022thin} & CVPR'22 & Thin-plate-spline & VoxCeleb1, CelebV, MUG & Image-to-video & ~100K videos & Learns non-rigid deformations via TPS, improves generalization in one-shot scenarios, and enhances motion realism using dense warping \\

HyperReenact \cite{bounareli2023hyperreenact} & ICCV'23 & StyleGAN-2 & VoxCeleb1, HDTF, CelebV-HQ & Video-to-video & ~100K videos & Employs hypernetworks and 3DMM controls for identity-preserving and expressive reenactment, robust to pose and lighting variation \\

FaceFusion \cite{facefusion2024} & GitHub'22 & GAN & Based on Wav2Lip, InsightFace & Video-to-video & N/A & Combines face swapping and lip-syncing in a modular real-time system, preserves identity and expression using multi-scale blending \\

FADM \cite{zeng2023face} & CVPR'23 & Diffusion & VoxCeleb2, CelebV, HDTF & Image-to-video & 100K+ (VoxCeleb), 1M (VoxCeleb2), 200K (CelebA) & Uses attention-based decomposition of expression and pose, enables accurate and high-quality face animation across multiple datasets \\

DreamTalk \cite{ma2023dreamtalk} & arXiv'23 & Diffusion & VoxCeleb2, HDTF, MEAD & Audio-to-video & 280K (MEAD, HDTF), 1M (VoxCeleb2) & Two-stage diffusion pipeline with multi-modal conditioning, excels in audio-lip-text alignment and photorealistic facial generation \\

AniFaceDiff \cite{chen2024anifacediff} & arXiv'24 & Diffusion & VoxCeleb2, TEDXPeople, MUG, CelebV-HQ & Audio-to-video & 1M videos (VoxCeleb2) & Dual-path diffusion transformer for expressive, identity-consistent talking-heads, supports zero-shot generation with high realism \\

VASA-1 \cite{xu2024vasa} & NeurIPS'24 & Diffusion Transformers & VASA, LRW, LRS3, CelebV & Audio-to-video & 1M (VoxCeleb2), ~22K (VASA) & Offers real-time and zero-shot talking-head generation with gaze and emotion control, driven by latent diffusion and transformer models \\

Diffused Heads \cite{stypulkowski2024diffused} & WACV'24 & Diffusion & VoxCeleb2, LRW, HDTF, MEAD & Audio-to-video & 100K+ (LRW) & Uses 3D face priors with dual-branch diffusion to decouple geometry and texture, enabling high control over expressions and identity \\

SyncTalk \cite{peng2024synctalk} & CVPR'24 & NeRF & VoxCeleb2, LRS2, CelebV-HQ & Audio-to-video & ~1M videos, 8K+ frames/video & Introduces sync-aware loss and multimodal attention, achieves superior lip synchronization and visual quality, suitable for real-time generation \\ \hline

\end{tabular}
\label{tab:merged_generation_detailed}
\end{table*}

\subsection{Deepfake Synthesis}

Deepfake synthesis has progressed rapidly with advances in generative modeling, particularly GANs, diffusion models, and NeRF-based frameworks, aimed at improving visual realism, identity preservation, and expression fidelity. Early methods like FaceVid \cite{wang2021one} and TPSMM \cite{zhao2022thin} introduced GAN-based and motion warping techniques for talking-head generation, leveraging keypoint decomposition and thin-plate spline motion estimation to animate source faces using driving signals. StyleGAN-based HyperReenact \cite{bounareli2023hyperreenact} enabled cross-subject reenactment by mapping facial features into latent space using hypernetworks, while FaceFusion \cite{facefusion2024} provided a real-time, modular system combining lip-syncing and face swapping. Diffusion-based approaches such as FADM \cite{zeng2023face}, AniFaceDiff \cite{chen2024anifacediff}, and DreamTalk \cite{ma2023dreamtalk} pushed the envelope by introducing attribute-guided conditioning, refined expression preservation, and emotion-aware modules for photorealistic and expressive video generation. VASA-1 \cite{xu2024vasa} combined diffusion and transformers to synthesize realistic facial dynamics and head movements in real time. Diffused Heads \cite{stypulkowski2024diffused} incorporated 3D priors and autoregressive sampling to model naturalistic behaviors, while SyncTalk \cite{peng2024synctalk}, the first NeRF-based method, featured synchronization-aware components for accurate lip motion, head pose, and hair-torso blending. These models were trained and evaluated on datasets such as VoxCeleb, HDTF, and CelebV, with recent works like SyncTalk, DreamTalk, Diffused Heads, and VASA-1 recognized as state-of-the-art due to their superior visual realism and temporal coherence. A summary of their architectures and characteristics is presented in Table~\ref{tab:merged_generation_detailed}, and representative visual outputs are illustrated in Figure~\ref{fig:synthesis-methods}.

\subsection{Deepfake Detection}
With the rapid advancement of deepfake generation techniques, the development of robust and generalizable detection models has become essential. Early approaches like MesoNet \cite{afchar2018mesonet} utilized lightweight CNN architectures (Meso-4 and MesoInception-4) to detect tampering based on facial features, while Xception \cite{rossler2019faceforensics++} introduced depthwise separable convolutions to efficiently learn spatial and temporal inconsistencies. Capsule networks \cite{nguyen2019capsule} further improved detection by modeling hierarchical spatial relationships, enhancing resistance to forgery across different manipulation types. EfficientNet \cite{tan2019efficientnet}, although not initially intended for deepfake detection, has been widely adopted due to its compound scaling strategy that yields high accuracy with fewer parameters. Benchmark datasets like FFD \cite{dang2020detection} and evaluation frameworks have been critical in standardizing assessments, providing diverse forgery types under varying compression levels. SRM \cite{luo2021generalizing} and RECCE \cite{cao2022end} introduced high-frequency noise analysis and reconstruction-based learning, respectively, to better identify subtle artifacts, while CORE \cite{ni2022core} focused on learning invariant features through representation consistency across data augmentations. UCF \cite{yan2023ucf} addressed generalization explicitly via a disentanglement framework, separating forgery-irrelevant, method-specific, and common forgery features through multi-task learning.
As summarized in Table~\ref{tab:summary_detection}, the field has shifted from early CNN-based classifiers to more sophisticated, representation-driven frameworks capable of generalizing across manipulation techniques and datasets, underscoring the arms race between synthesis and detection.

\subsection{Human Level Detection}

\begin{figure}[h]
    \centering
    \includegraphics[width=0.5\textwidth]{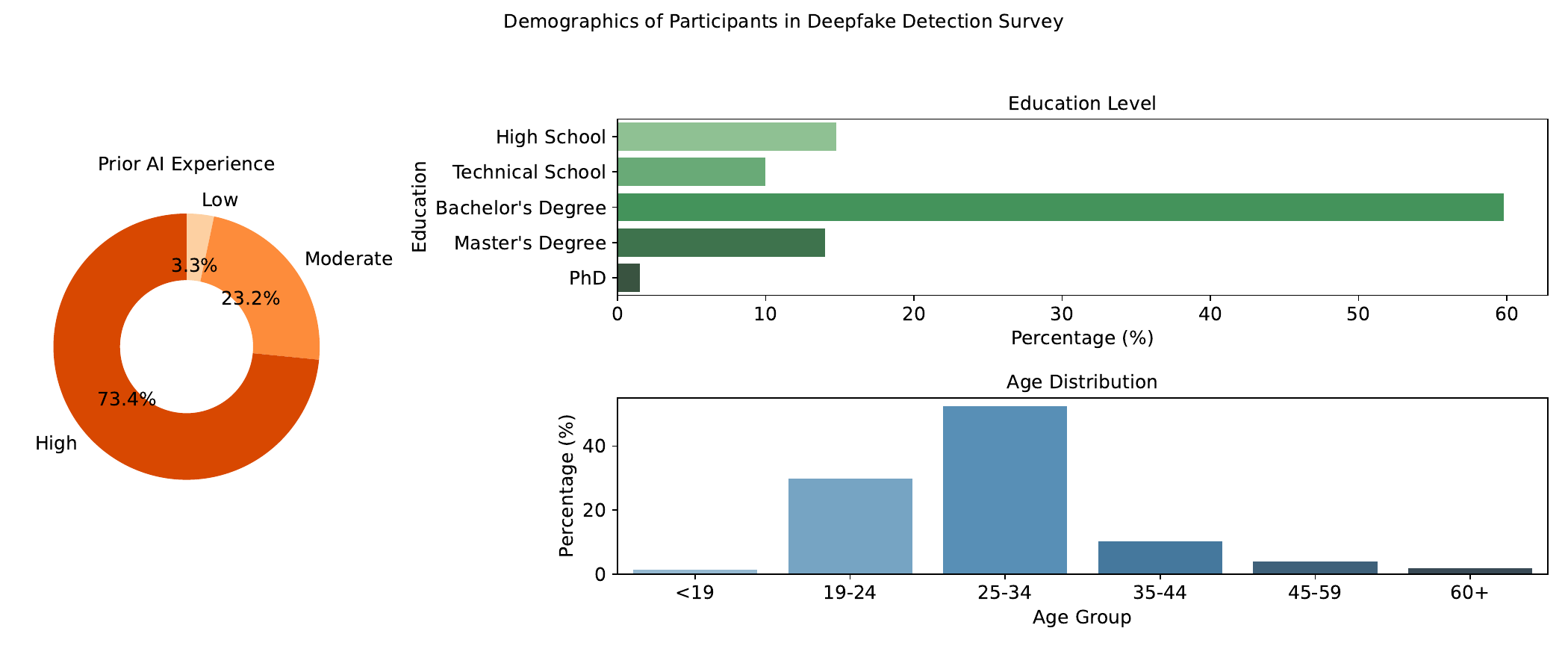}
    \caption{Demographic breakdown of participants in the deepfake detection survey. The majority of participants reported high prior AI experience (73.4\%) and held at least a bachelor’s degree, with over half possessing a master’s degree. The largest age group was 25–34 years, indicating a relatively young and technologically proficient participant pool.}
    \label{fig:human-demographics}
\end{figure}

To assess human ability to detect deepfakes, we conducted a survey using a structured questionnaire hosted on the Qualtrics platform. A total of 271 participants successfully completed the survey. Participants were recruited through Amazon Mechanical Turk and Prolific, with each participant being compensated SGD \$5 for their involvement. On average, participants spent approximately 20 minutes completing the survey. The demographic distribution of the participants is illustrated in Figure \ref{fig:human-demographics}.

The demographic data reveals a predominantly young, well-educated participant group, with 52.4\% of respondents aged 25-34 and 59.8\% holding a Bachelor's degree. This suggests that the survey attracted a tech-savvy audience, which may influence their ability to detect deepfakes. The inclusion of participants across a range of ages and educational backgrounds provides a broader perspective on the human detection capabilities of deepfake videos. Participants were also categorized based on their prior experience with AI-generating algorithms, reflecting varying levels of familiarity with advanced AI tools. The participants were divided into three clusters: Low Experience, Moderate Experience, and High Experience. Cluster 1, Low Experience, consists of participants who have never heard of AI-generating algorithms like chatGPT, GPT-4, DALL-E, MidJourney, or StableDiffusion, accounting for 3.4\% of the total participants. Cluster 2, Moderate Experience, includes those who have heard of chatGPT or GPT-4, but not DALL-E, MidJourney, or StableDiffusion, or have heard of at least one of these tools, comprising 23.2\% of participants. Cluster 3, High Experience, consists of participants who have used chatGPT/GPT-4 or DALL-E/MidJourney/StableDiffusion, or both, representing 73.4\% of the participants. This distribution highlights the prevalence of AI tool usage among participants, with a substantial majority having significant experience with these technologies.

\begin{landscape}
\begin{table}[]
\centering
\footnotesize
\def\arraystretch{1.2}
\caption{Summary of different deepfake detection methods, including their publication details, datasets, techniques, model types, and key highlights.}
\begin{tabular}{P{1.4cm}P{1.8cm}P{2.2cm}P{3.6cm}P{1cm}P{2.5cm}p{8cm}}
\hline
\textbf{Name} &
  \textbf{Publishing} &
  \textbf{Dataset trained on} &
  \textbf{Method} &
  \textbf{Type} &
 \textbf{Number of images/videos} &
  \textbf{Highlights} \\ \hline

MesoNet \cite{afchar2018mesonet} &
  WIFS 2018 &
  Deepfake, Face2Face \cite{thies2016face2face} &
  Incorporates modified inception modules with 3×3 dilated conv. and 1×1 conv and skip-connections, builds upon the Meso-4 architecture &
  Naive &
  1000+ videos, DeepFake \& Face2Face &
  Comprises four layers of convolutions and pooling for feature extraction, Achieves over 98\% detection rate for Deepfake videos, Contains 27,977 trainable parameters for efficient processing, Employs ReLU activation, Batch Normalization, and Dropout for improved generalization, Specifically designed to detect face tampering in videos \\ 


Xception \cite{rossler2019faceforensics++} &
  ICCV 2019 &
  ImageNet \cite{russakovsky2015imagenet}, fine tuned on FaceForensics++ \cite{rossler2019faceforensics++} &
  CNN architecture with separable convolutions and residual connections, pre-trained on the ImageNet \cite{russakovsky2015imagenet} dataset & 
  Naive &
  ImageNet: 14m images, FaceForensics++: 1000 real, 4000 fake videos &
  Outperforms other architectures in face forgery detection, Utilizes pre-trained weights from ImageNet for better generalization, Final layer modified for two outputs specific to forgery detection, Pre-trained for 3 epochs, then trained for 15 more epochs, Maintains performance on compressed videos \\
  
Capsule \cite{nguyen2019capsule} &
  ICASSP 2019 &
  FaceForensics \cite{rossler2018faceforensics} &
  Capsule networks, Dynamic routing algorithm (VGG-19 network) \cite{simonyan2014very} for feature extraction &
  Spatial &
  $\sim$1000 videos &
  Lower half total error rate than state-of-the-art methods, Improved accuracy using capsule network with VGG-19 features, Perfect accuracy in distinguishing CGI from photographic images, Robustness against adversarial attacks via training with random noise, Validation through comprehensive experiments on forgery detection \\ 

EfficientNet-B4 \cite{tan2019efficientnet} &
  ICML 2019 &
  ImageNet \cite{russakovsky2015imagenet} &
  Compound scaling for uniform adjustment of network dimensions, EfficientNet, a mobile-size baseline network optimized for accuracy and FLOPS &
  Naive &
  $\sim$1.2 million images &
  Introduces a new family of models that optimize accuracy and efficiency through compound scaling; EfficientNet Architecture, Achieves top results on various benchmarks, including ImageNet, with fewer parameters, Proposes a systematic approach to scaling networks in width, depth, and resolution, Demonstrates effectiveness in transfer learning across multiple datasets and tasks, Significantly reduces computational costs while maintaining high accuracy \\ 
  
FFD \cite{dang2020detection} &
  CVPR 2020 &
  Diverse Fake Face Dataset \cite{dang2020detection} &
  CNN-based network with an attention mechanism, XceptionNet for depthwise separable convolutions and residual connections &
  Spatial &
  2.6 million images &
  Enhances localization of manipulated regions for improved detection accuracy, Achieves AUC of 99.64 with low EER values using XceptionNet, Tested against benchmarks, outperforming existing methods in facial forgery detection, Utilizes both XceptionNet and VGG16 for comprehensive performance analysis \\ 
  
SRM \cite{luo2021generalizing} &
  CVPR 2021 &
  FaceForensics++ \cite{rossler2019faceforensics++} &
  Two-stream network architecture, residual guided spatial and dual cross-modality attention modules, Xception backbone &
  Frequency &
  1000 real, 4000 fake videos &
  Two-stream network for face forgery detection, Residual guided spatial and dual cross-modality attention modules, Outperforms existing methods on multiple datasets, Validates effectiveness through extensive testing, Acknowledges backing from research programs and funding bodies \\ 
  
RECCE \cite{cao2022end} &
  CVPR 2022 &
  FaceForensics++, Celeb-DF, WildDeepfake, DFDC &
  Reconstruction learning, multi-scale graph reasoning, and reconstruction guided attention &
  Spatial &
  Around 150K videos accross datasets &
  Improves accuracy in detecting face forgery, Handles unknown forgery patterns effectively, Analyzes discrepancies at different levels, Focuses on critical forgery traces for better classification, Outperforms existing methods on benchmark datasets \\ 

CORE \cite{ni2022core} &
  CVPR 2022 &
  FaceForensics++ \cite{rossler2019faceforensics++} &
  Consistent Representation Learning, Consistency loss with different penalties, Xception network \cite{chollet2017xception} for face forgery detection &
  Spatial &
  1000 real, 4000 fake videos &
  CORE (Consistent Representation Learning) framework is introduced for effective face forgery detection, Achieves state-of-the-art performance on the FF++ and Celeb-DF datasets. Demonstrates significant gains over existing methods, particularly in cross-dataset evaluations, Integrates flexibly with various backbone networks, showcasing its versatility \\ 
 
UCF \cite{yan2023ucf} &
  CVPR 2023 &
  FaceForensics++ \cite{rossler2019faceforensics++} &
  Disentanglement framework to separate content and forgery features, Multi-task learning strategy to enhance detection performance, Primarily utilizes Xception architecture &
  Spatial &
  1,000 pristine videos &
  Superior performance across 27 detectors, reducing overfitting to specific artifacts, Separates content and fingerprint features using multi-task learning for enhanced detection, Achieves top AUC scores of 0.824 and 0.805 on CelebDF and DFDC datasets, Compatible with various architectures like Xception, ConvNeXt, and ResNet, Extensive ablation studies demonstrate improved generalization and resilience to biases \\ \hline
\end{tabular}%
\\
\label{tab:summary_detection}
\end{table}
\end{landscape}

Participants viewed 20 five-second, silent video clips selected from a pool of 320 stimuli. These included 8 categories—7 deepfake types and 1 real—each with 20 unique videos presented in both low- and high-resolution formats, totaling 320 videos $(8 \times 20 \times 2)$.

This study assessed participants' ability to detect deepfakes across various video types and resolutions using an anonymous, IRB-approved survey, with informed consent obtained from all participants. The design enabled a comprehensive evaluation of detection performance across manipulation categories and quality levels. For model evaluation, we employed a wide range of metrics: AUC and AP to assess overall classification performance, Precision and Recall to measure accuracy in identifying true positives and negatives, and \$d'\$ and C from signal detection theory to analyze discriminability and response bias. This robust framework supports a detailed comparison of both human and algorithmic detection capabilities.

\section{Results}

\subsection{Generating synthetic videos}
We used open-source models from original authors to generate videos with source images and driving videos from the VoxCeleb dataset, ensuring transparency, reproducibility, and use of state-of-the-art techniques. Selection was based on video quality, model/code availability, and publication date—prioritizing high-quality synthesis for accurate evaluation, consistent implementation, and relevance to recent advancements. This curated set supported robust comparisons across both algorithmic and human detection.

\subsection{Deepfake Detection}

This section presents a comparative analysis of deepfake detection performance between automated algorithms and human participants. Key evaluation metrics include the Area Under the ROC Curve (AUC) and Average Precision (AP), which assess a model’s ability to distinguish real from fake videos using prediction probabilities.

Since humans do not provide probabilistic outputs, participants rated each video’s authenticity on a five-point scale ranging from “Definitely computer generated” to “Definitely taken by camera.” These responses were converted to probability scores, with values from 1.00 to 0.00, and averaged per video to enable comparison with model predictions.

Results are summarized in Tables \ref{tab:comparison_table_auc_ap}, \ref{tab:comparison_table_precision_recall}, and \ref{tab:comparison_table_dprime_C}, covering AUC, AP, Precision, Recall, and signal detection theory metrics such as d-prime and criterion, along with the mean and standard deviation across models and manipulation types.

Human evaluators outperformed automated methods, achieving a mean AUC of 93.10 and a mean AP of 94.81, with low standard deviations that reflect high accuracy and consistency. Among models, FFD \cite{dang2020detection} and UCF \cite{yan2023ucf} showed relatively strong performance. For instance, FFD reached an AUC of 69.38 and an AP of 71.00. In contrast, MesoNet \cite{afchar2018mesonet} demonstrated much higher variability, signaling inconsistent detection performance.

Table \ref{tab:comparison_table_precision_recall} further highlights the robustness of human detection. Human evaluators showed a high mean precision of 98.93 and a moderate recall of 70.72. While recall was more variable, the consistently high precision suggests that humans tend to be cautious yet accurate. Capsule \cite{nguyen2019capsule} and RECCE \cite{cao2022end} achieved competitive precision but suffered from elevated variability in recall. MesoNet’s uniformly perfect precision and recall scores of 100 percent appear anomalous and may indicate artifacts or thresholding issues that warrant further examination.

Table \ref{tab:comparison_table_dprime_C} frames detection performance using signal detection theory. Human participants achieved a high mean d-prime of 7.01 with low variability, reflecting strong and consistent discrimination between real and fake videos. Their average criterion of 2.86 suggests a cautious response bias, aligning with their high precision. Automated models showed much weaker discrimination, with CORE \cite{ni2022core} and EfficientNet-B4 \cite{tan2019efficientnet} scoring d-primes below 1 and greater variability overall. MesoNet \cite{afchar2018mesonet} again displayed unreliable results, with a d-prime of 0 and an implausibly extreme criterion of -6.36.

\begin{figure}[t]
    \centering
    \includegraphics[width=0.5\textwidth]{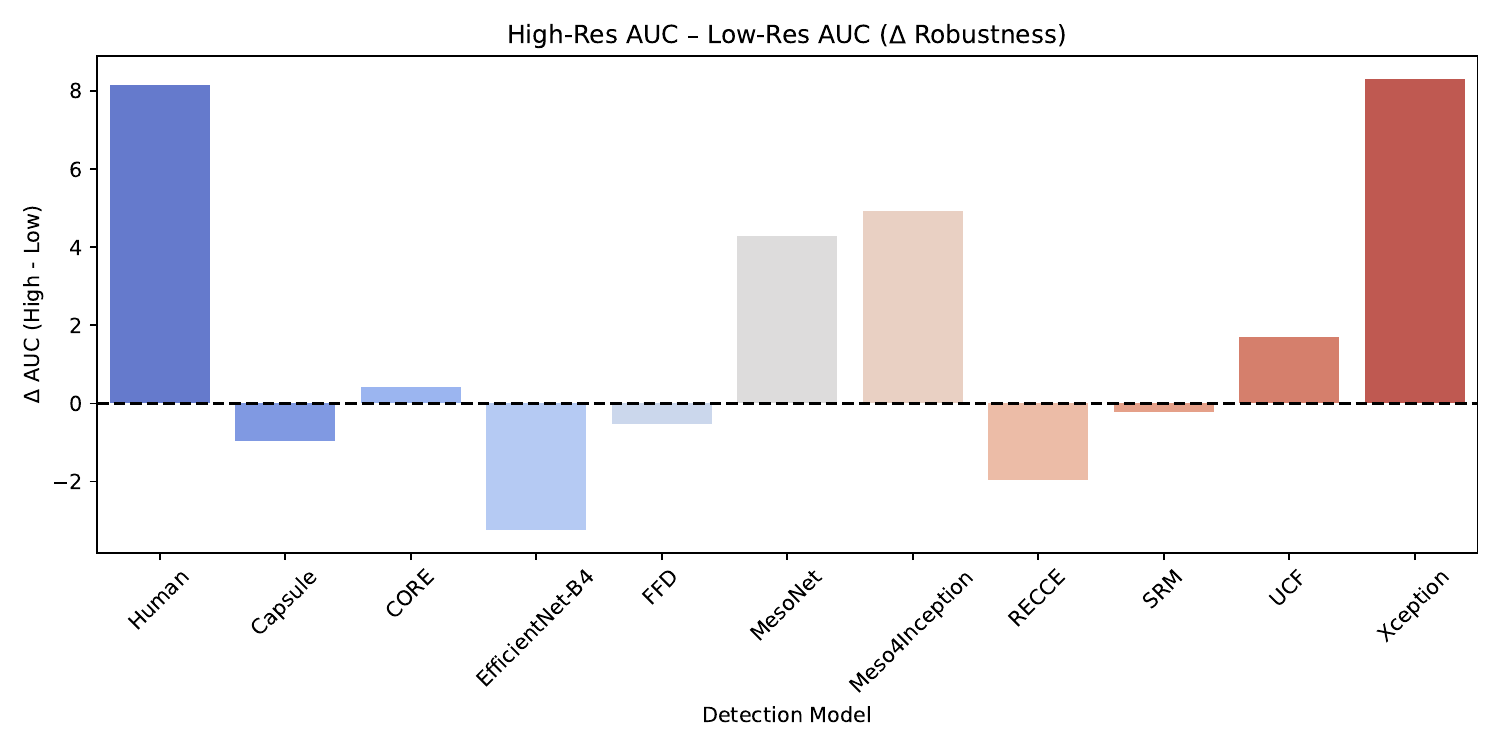}
    \caption{$\Delta \text{AUC}$ (High-Res -- Low-Res) for each detection model, representing robustness to video resolution changes. Each AUC value is the average performance across multiple generative models (e.g., different deepfake sources). Positive $\Delta$ values indicate better performance on high-resolution videos. Xception \cite{rossler2019faceforensics++} and Human detectors show the largest gains ($>$8 AUC points), while EfficientNet-B4 \cite{tan2019efficientnet} and RECCE \cite{cao2022end} perform worse on high-res inputs. The overall average $\Delta$AUC across all models is +1.90, suggesting that high-resolution videos modestly improve detection performance on average.}
    \label{fig:delta_auc}
\end{figure}

\begin{landscape}

\begin{table}[h!]
\centering
\def\arraystretch{1.2}
\caption{Comparison of Detection and Generation Performance (AUC and AP values) on 5-Second Videos for Various Deepfake Detection Models (ROWS) and Generators (COLUMNS). The bolded values highlight the best overall detection model across rows and the most challenging deepfake generator across columns.}
\resizebox{1.35\textwidth}{!}{%
    \begin{tabular}{lcccccccccccccccc}
        \hline
        \multirow{2}{*}{Model} &
        \multicolumn{2}{c}{AniFaceDiff \cite{chen2024anifacediff}} &
        \multicolumn{2}{c}{TPSMM \cite{zhao2022thin}} &
        \multicolumn{2}{c}{FaceVid \cite{wang2021one}} &
        \multicolumn{2}{c}{HyperReenact \cite{bounareli2023hyperreenact}} &
        \multicolumn{2}{c}{FaceFusion \cite{facefusion2024}} &
        \multicolumn{2}{c}{FADM \cite{zeng2023face}} &
        \multicolumn{2}{c}{SOTA \cite{xu2024vasa, ma2023dreamtalk, stypulkowski2024diffused, peng2024synctalk}} &
        \multicolumn{2}{c} {Detection Performance: Mean $\pm$ Std}\\ 
         & AUC & AP & AUC & AP & AUC & AP & AUC & AP & AUC & AP & AUC & AP & AUC & AP & AUC & AP \\ \hline
        Human & 99.38 & 99.43 & 90.44 & 93.85 & 89.20 & 91.71 & 98.38 & 98.92 & 92.75 & 94.65 & 90.16 & 92.67 & 91.41 & 92.47 & \textbf{93.10} $\pm$ 4.11 & \textbf{94.81} $\pm$ 3.13 \\
        Capsule \cite{nguyen2019capsule} & 49.94 & 49.70 & 71.31 & 72.74 & 78.56 & 77.09 & 69.19 & 65.98 & 65.56 & 61.67 & 83.94 & 81.6 & 47.38 & 51.66 & 66.55 $\pm$ 13.67 & 65.78 $\pm$ 12.26 \\
        CORE \cite{ni2022core} & 50.44 & 53.86 & 76.75 & 75.43 & 69.25 & 70.51 & 80.19 & 81.38 & 72.56 & 74.56 & 79.62 & 79.05 & 52.37 & 60.60 & 68.74 $\pm$ 12.46 & 70.77 $\pm$ 10.06 \\
        EfficientNet-B4 \cite{tan2019efficientnet} & 50.19 & 55.03 & 63.06 & 69.97 & 71.06 & 74.23 & 68.56 & 70.64 & 67.44 & 72.84 & 78.00 & 79.92 & 51.37 & 66.00 & 64.24 $\pm$ 10.24 & 69.80 $\pm$ 7.80 \\
        FFD \cite{dang2020detection} & 53.69 & 54.57 & 65.62 & 67.21 & 66.69 & 71.34 & 69.44 & 71.47 & 79.38 & 80.17 & 79.81 & 81.66 & 71.00 & 70.58 & 69.38 $\pm$ 8.93 & 71.00 $\pm$ 8.97 \\
        MesoNet \cite{afchar2018mesonet} & 40.88 & 48.44 & 32.88 & 39.07 & 42.19 & 45.43 & 83.00 & 84.43 & 47.06 & 52.87 & 39.50 & 48.00 & 62.62 & 66.85 & 49.73 $\pm$ 17.34 & 55.01 $\pm$ 15.54 \\
        Meso4Inception \cite{afchar2018mesonet} & 53.62 & 53.15 & 60.88 & 58.77 & 69.12 & 69.08 & 57.00 & 54.82 & 63.62 & 61.16 & 73.88 & 67.71 & 55.75 & 56.71 & 61.98 $\pm$ 7.42 & 60.20 $\pm$ 6.18 \\
        RECCE \cite{cao2022end} & 45.31 & 52.33 & 79.00 & 82.95 & 69.69 & 74.86 & 85.81 & 88.42 & 81.12 & 85.93 & 84.25 & 86.83 & 51.38 & 63.57 & 70.94 $\pm$ 16.37 & 76.41 $\pm$ 13.74 \\
        SRM \cite{luo2021generalizing} & 43.50 & 47.36 & 66.75 & 64.64 & 63.87 & 62.69 & 70.19 & 65.75 & 73.81 & 75.49 & 76.88 & 73.08 & 49.00 & 58.72 & 63.43 $\pm$ 12.59 & 63.96 $\pm$ 9.35 \\
        UCF \cite{yan2023ucf} & 56.12 & 57.49 & 75.38 & 75.83 & 71.12 & 74.38 & 77.50 & 75.71 & 78.19 & 78.89 & 81.12 & 77.43 & 68.00 & 71.67 & 72.49 $\pm$ 8.47 & 73.06 $\pm$ 7.23 \\
        Xception \cite{rossler2019faceforensics++} & 43.62 & 45.07 & 66.12 & 66.58 & 69.31 & 65.30 & 74.06 & 67.68 & 76.25 & 78.12 & 81.12 & 77.54 & 68.50 & 66.20 & 68.43 $\pm$ 12.08 & 66.64 $\pm$ 10.95 \\
        Generation Performance: Mean $\pm$ Std & \textbf{53.34} $\pm$ 16.03 & \textbf{56.04} $\pm$ 14.86 & 68.02 $\pm$ 14.44 & 69.73 $\pm$ 13.92 & 69.10 $\pm$ 11.22 & 70.60 $\pm$ 11.23 & 75.76 $\pm$ 10.99 & 75.02 $\pm$ 12.42 & 72.52 $\pm$ 11.75 & 74.21 $\pm$ 11.89 & 77.12 $\pm$ 13.19 & 76.86 $\pm$ 11.58 & 60.80 $\pm$ 13.23 & 65.91 $\pm$ 10.65 \\ \hline 
    \end{tabular}%
}
\label{tab:comparison_table_auc_ap}
\end{table}

\begin{table}[h!]
\centering
\def\arraystretch{1.2}
\caption{Comparison of Detection and Generation Performance (P, Precision and R, Recall Values) on 5-Second Videos for Various Deepfake Detection Models (ROWS) and Generators (COLUMNS). The bolded values highlight the best overall detection model across rows and the most challenging deepfake generator across columns.}
\resizebox{1.35\textwidth}{!}{%
\begin{tabular}{lcccccccccccccccc}
    \hline
    \multirow{2}{*}{Model} &
    \multicolumn{2}{c}{AniFaceDiff \cite{chen2024anifacediff}} &
    \multicolumn{2}{c}{TPSMM \cite{zhao2022thin}} &
    \multicolumn{2}{c}{FaceVid \cite{wang2021one}} &
    \multicolumn{2}{c}{HyperReenact \cite{bounareli2023hyperreenact}} &
    \multicolumn{2}{c}{FaceFusion \cite{facefusion2024}} &
    \multicolumn{2}{c}{FADM \cite{zeng2023face}} &
    \multicolumn{2}{c}{SOTA \cite{xu2024vasa, ma2023dreamtalk, stypulkowski2024diffused, peng2024synctalk}} &
    \multicolumn{2}{c} {Detection Performance: Mean ± Std}\\ 
    & P & R & P & R & P & R & P & R & P & R & P & R & P & R & P & R \\ \hline

    Human & 92.50 & 96.10 & 100.00 & 67.50 & 100.00 & 58.97 & 100.00 & 95.00 & 100.00 & 75.00 & 100.00 & 55.00 & 100.00 & 47.50 & \textbf{98.93} $\pm$ 2.83 & 70.72 $\pm$ 19.09 \\
    Capsule \cite{nguyen2019capsule} & 50.00 & 47.50 & 61.22 & 75.00 & 64.81 & 87.50 & 60.42 & 72.50 & 59.57 & 70.00 & 66.07 & 92.50 & 42.42 & 35.00 & 57.79 $\pm$ 8.53 & 68.57 $\pm$ 20.66 \\
    CORE \cite{ni2022core} & 43.75 & 17.50 & 67.86 & 47.50 & 66.67 & 45.00 & 70.97 & 55.00 & 70.00 & 52.50 & 73.53 & 62.50 & 60.87 & 35.00 & 64.81 $\pm$ 10.11 & 45.00 $\pm$ 14.86 \\
    EfficientNet-B4 \cite{tan2019efficientnet} & 52.63 & 25.00 & 65.38 & 42.50 & 67.86 & 47.50 & 64.00 & 40.00 & 66.67 & 45.00 & 70.00 & 52.50 & 60.87 & 35.00 & 63.92 $\pm$ 5.76 & 41.07 $\pm$ 9.00 \\
    FFD \cite{dang2020detection} & 52.63 & 25.00 & 64.00 & 40.00 & 66.67 & 45.00 & 66.67 & 45.00 & 71.88 & 57.50 & 74.29 & 65.00 & 66.67 & 45.00 & 66.12 $\pm$ 6.92 & 46.07 $\pm$ 12.74 \\
    MesoNet \cite{afchar2018mesonet} & 50.00 & 100.00 & 50.00 & 100.00 & 50.00 & 100.00 & 50.00 & 100.00 & 50.00 & 100.00 & 50.00 & 100.00 & 50.00 & 100.00 & 50.00 $\pm$ 0.00 & \textbf{100.00} $\pm$ 0.00 \\
    Meso4Inception \cite{afchar2018mesonet} & 50.91 & 70.00 & 55.74 & 85.00 & 57.14 & 90.00 & 53.45 & 77.50 & 55.00 & 82.50 & 59.09 & 97.50 & 54.24 & 80.00 & 55.08 $\pm$ 2.63 & 83.21 $\pm$ 8.86 \\
    RECCE \cite{cao2022end} & 62.50 & 12.50 & 86.96 & 50.00 & 81.25 & 32.50 & 88.00 & 55.00 & 88.46 & 57.50 & 88.00 & 55.00 & 76.92 & 25.00 & 81.73 $\pm$ 9.53 & 41.07 $\pm$ 17.73 \\
    SRM \cite{luo2021generalizing} & 46.43 & 32.50 & 63.41 & 65.00 & 59.46 & 55.00 & 65.91 & 72.50 & 66.67 & 75.00 & 67.39 & 77.50 & 48.28 & 35.00 & 59.65 $\pm$ 8.82 & 58.93 $\pm$ 18.76 \\
    UCF \cite{yan2023ucf} & 50.00 & 12.50 & 73.68 & 35.00 & 73.68 & 35.00 & 66.67 & 25.00 & 75.00 & 37.50 & 76.19 & 40.00 & 66.67 & 25.00 & 68.84 $\pm$ 9.16 & 30.00 $\pm$ 9.68 \\
    Xception \cite{rossler2019faceforensics++} & 38.89 & 17.50 & 68.57 & 60.00 & 69.44 & 62.50 & 66.67 & 55.00 & 72.50 & 72.50 & 73.81 & 77.50 & 66.67 & 55.00 & 65.22 $\pm$ 11.93 & 57.14 $\pm$ 19.44 \\
    Generation Performance: Mean ± Std & \textbf{49.77} $\pm$ 6.20 & \textbf{36.00} $\pm$ 28.73 & 65.68 $\pm$ 10.01 & 60.00 $\pm$ 21.25 & 65.70 $\pm$ 8.73 & 60.00 $\pm$ 24.21 & 65.28 $\pm$ 10.29 & 59.75 $\pm$ 21.46 & 67.58 $\pm$ 10.92 & 65.00 $\pm$ 18.67 & 69.84 $\pm$ 10.27 & 72.00 $\pm$ 20.41 & 59.36 $\pm$ 10.52 & 47.00 $\pm$ 24.74 \\ 

    \hline

\end{tabular}%
}

\label{tab:comparison_table_precision_recall}
\end{table}

\begin{table}[h!]
\centering
\def\arraystretch{1.2}
\caption{Comparison of Detection and Generation Performance ($d\text{'}$, Sensitivity Index and C, Decision Criterion Values) on 5-Second Videos for Various Deepfake Detection Models (ROWS) and Generators (COLUMNS). The bolded values highlight the best overall detection model across rows and the most challenging deepfake generator across columns.}
\resizebox{1.35\textwidth}{!}{%
\begin{tabular}{lcccccccccccccccc}
\hline
\multirow{2}{*}{Model} &
  \multicolumn{2}{c}{AniFaceDiff \cite{chen2024anifacediff}} &
  \multicolumn{2}{c}{TPSMM \cite{zhao2022thin}} &
  \multicolumn{2}{c}{FaceVid \cite{wang2021one}} &
  \multicolumn{2}{c}{HyperReenact \cite{bounareli2023hyperreenact}} &
  \multicolumn{2}{c}{FaceFusion \cite{facefusion2024}} &
  \multicolumn{2}{c}{FADM \cite{zeng2023face}} &
  \multicolumn{2}{c}{SOTA \cite{xu2024vasa, ma2023dreamtalk, stypulkowski2024diffused, peng2024synctalk}} &
  \multicolumn{2}{c} {Detection Performance: Mean ± Std}\\ 
 & d' & C & d' & C & d' & C & d' & C & d' & C & d' & C & d' & C & d' & C \\ \hline
Human & 7.80 & 2.46 & 6.82 & 2.95 & 6.59 & 3.07 & 8.01 & 2.36 & 7.04 & 2.84 & 6.49 & 3.12 & 6.30 & 3.21 & \textbf{7.01} $\pm$ 0.66 & \textbf{2.86} $\pm$ 0.33 \\
Capsule \cite{nguyen2019capsule} & 0.00 & 0.06 & 0.74 & -0.31 & 1.21 & -0.54 & 0.66 & -0.27 & 0.59 & -0.23 & 1.50 & -0.69 & -0.32 & 0.22 & 0.63 $\pm$ 0.63 & -0.25 $\pm$ 0.32 \\
CORE \cite{ni2022core} & -0.18 & 0.85 & 0.69 & 0.41 & 0.63 & 0.44 & 0.88 & 0.31 & 0.82 & 0.35 & 1.07 & 0.22 & 0.37 & 0.57 & 0.61 $\pm$ 0.41 & 0.45 $\pm$ 0.21 \\
EfficientNet-B4 \cite{tan2019efficientnet} & 0.08 & 0.71 & 0.57 & 0.47 & 0.69 & 0.41 & 0.50 & 0.50 & 0.63 & 0.44 & 0.82 & 0.35 & 0.37 & 0.57 & 0.52 $\pm$ 0.24 & 0.49 $\pm$ 0.12 \\
FFD \cite{dang2020detection} & 0.08 & 0.71 & 0.50 & 0.50 & 0.63 & 0.44 & 0.63 & 0.44 & 0.94 & 0.28 & 1.14 & 0.19 & 0.63 & 0.44 & 0.65 $\pm$ 0.33 & 0.43 $\pm$ 0.17 \\
MesoNet \cite{afchar2018mesonet} & 0.00 & -6.36 & 0.00 & -6.36 & 0.00 & -6.36 & 0.00 & -6.36 & 0.00 & -6.36 & 0.00 & -6.36 & 0.00 & -6.36 & 0.00 $\pm$ 0.00 & -6.36 $\pm$ 0.00 \\
Meso4Inception \cite{afchar2018mesonet} & 0.07 & -0.49 & 0.58 & -0.75 & 0.83 & -0.87 & 0.30 & -0.60 & 0.48 & -0.69 & 1.51 & -1.21 & 0.39 & -0.65 & 0.59 $\pm$ 0.47 & -0.75 $\pm$ 0.23 \\
RECCE \cite{cao2022end} & 0.29 & 1.29 & 1.44 & 0.72 & 0.99 & 0.95 & 1.57 & 0.66 & 1.63 & 0.63 & 1.57 & 0.66 & 0.77 & 1.06 & 1.18 $\pm$ 0.51 & 0.85 $\pm$ 0.25 \\
SRM \cite{luo2021generalizing} & -0.14 & 0.39 & 0.70 & -0.03 & 0.44 & 0.10 & 0.92 & -0.14 & 0.99 & -0.18 & 1.07 & -0.22 & -0.07 & 0.35 & 0.56 $\pm$ 0.5 & 0.04 $\pm$ 0.25 \\
UCF \cite{yan2023ucf} & 0.00 & 1.15 & 0.77 & 0.77 & 0.77 & 0.77 & 0.48 & 0.91 & 0.83 & 0.73 & 0.90 & 0.70 & 0.48 & 0.91 & 0.60 $\pm$ 0.31 & 0.85 $\pm$ 0.16 \\
Xception \cite{rossler2019faceforensics++} & -0.34 & 0.77 & 0.85 & 0.17 & 0.92 & 0.14 & 0.72 & 0.24 & 1.20 & 0.00 & 1.35 & -0.08 & 0.72 & 0.24 & 0.77 $\pm$ 0.55 & 0.21 $\pm$ 0.27 \\
Generation Performance: Mean ± Std & \textbf{-0.01} $\pm$ 0.17 & -0.09 $\pm$ 2.26 & 0.68 $\pm$ 0.36 & -0.44 $\pm$ 2.13 & 0.71 $\pm$ 0.33 & -0.45 $\pm$ 2.15 & 0.67 $\pm$ 0.42 & -0.43 $\pm$ 2.13 & 0.81 $\pm$ 0.44 & -0.50 $\pm$ 2.10 & 1.09 $\pm$ 0.46 & \textbf{-0.64} $\pm$ 2.09 & 0.33 $\pm$ 0.36 & -0.27 $\pm$ 2.19\\ \hline 
\end{tabular}%
}

\label{tab:comparison_table_dprime_C}
\end{table}

\end{landscape}
To evaluate robustness to resolution, we analyzed detection performance on high- versus low-resolution videos. As shown in Figure \ref{fig:delta_auc}, we computed the change in AUC ($\Delta \text{AUC}$) for each model. Human and Xception \cite{rossler2019faceforensics++} detectors showed the largest improvements, both gaining over 8 AUC points on high-resolution inputs. Meanwhile, EfficientNet-B4 and RECCE \cite{cao2022end} performed worse on high-res data, hinting at overfitting or limited generalization. On average, detection models showed a modest improvement of 1.90 AUC points, emphasizing that video quality can significantly influence deepfake detection effectiveness.

\begin{figure}
    \centering
    \includegraphics[width=0.5\textwidth]{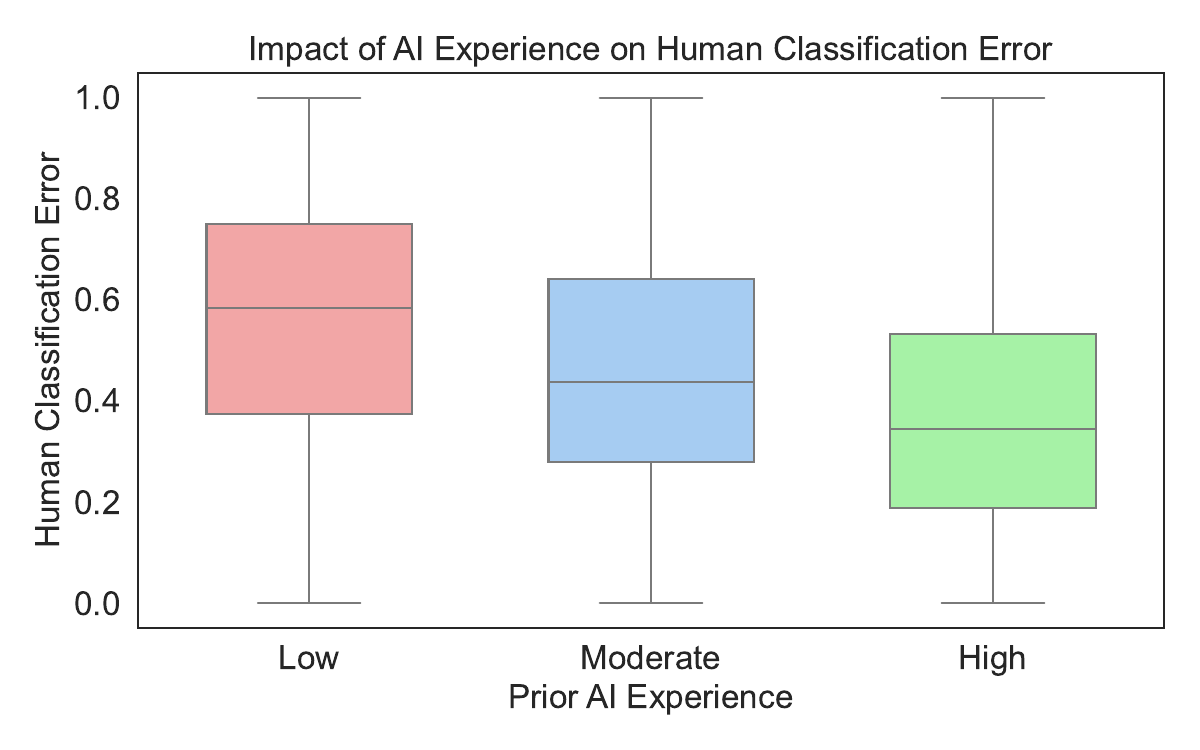}
    \caption{Aggregate impact of prior AI experience on fake video classification accuracy.
This figure shows the average classification error across participant groups with different levels of AI experience. Participants with low AI experience had the highest overall error rates, while those with high experience performed significantly better. These results indicate a clear negative correlation between AI expertise and misclassification of fake videos at the group level.}
    \label{fig:classification_error_vs_ai_experience}
\end{figure}

\begin{figure}
    \centering
    \includegraphics[width=0.5\textwidth]{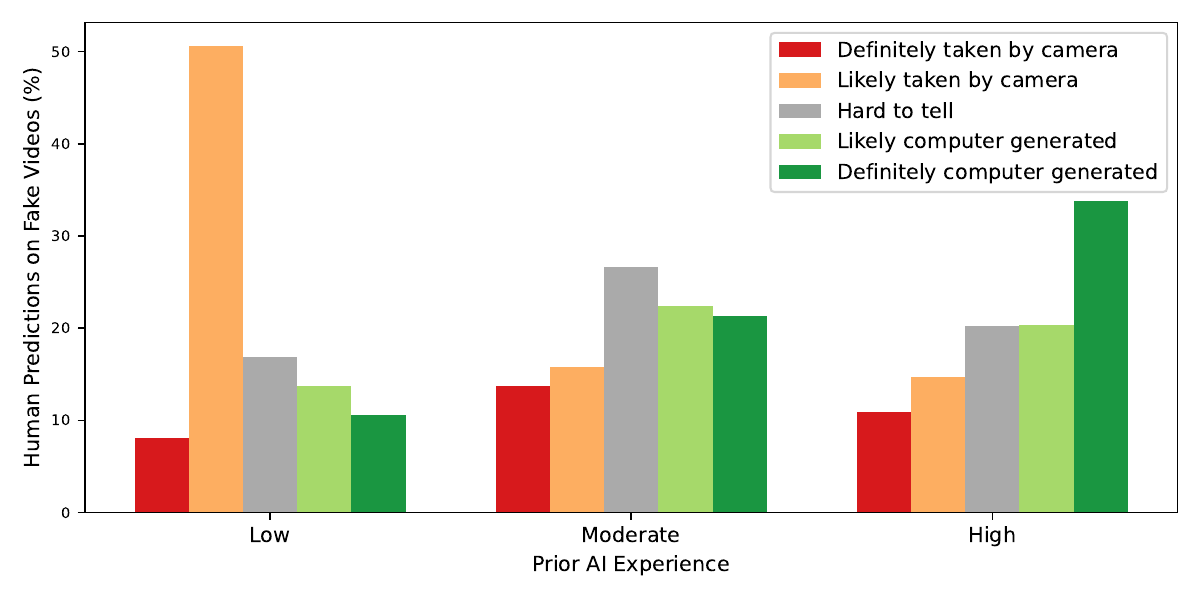}
    \caption{Individual classification responses to fake videos by AI experience level.
This figure captures how individual participants rated individual fake videos, broken down by their level of prior AI experience. Participants with low AI experience were more likely to misclassify fake content as real, whereas those with high experience more frequently labeled videos as ‘Definitely computer generated.’ The results suggest that AI expertise improves detection accuracy on a video-by-video basis.}
    \label{fig:human_predictions_fake_videos}
\end{figure}

We analyzed how participants' prior experience with generative AI tools influenced their ability to detect deepfakes. Based on self-reported familiarity and usage of tools like ChatGPT, DALL·E, and MidJourney, participants were grouped into low, moderate, and high experience levels. As shown in Figure \ref{fig:classification_error_vs_ai_experience}, detection error declined with increasing AI experience, with high-experience participants achieving the lowest error rates. A one-way ANOVA confirmed a significant effect of AI experience on discriminability (p = 0.0137), highlighting the value of hands-on exposure in improving media forensics skills. Further, as shown in Figure \ref{fig:human_predictions_fake_videos}, low-experience participants tended to misclassify deepfakes as real, while high-experience individuals were more confident and accurate in identifying manipulated content. These findings suggest that familiarity with generative AI enhances both accuracy and confidence in synthetic media detection.


Diffusion-based deepfakes are inherently more challenging to detect than GAN-based ones due to differences in both their generative objectives and loss formulations. GANs optimize a minimax adversarial loss, often yielding visually sharp outputs but with distinctive, generator-specific artifacts, whereas diffusion models minimize a denoising score-matching loss that iteratively refines samples to better match real data statistics, reducing detectable cues. Measured via FID and FVD \cite{chen2024anifacediff}, diffusion content lies significantly closer to real distributions, leading to higher Bayes error and lower detectability; moreover, detectors trained on GAN fakes overfit to generator-specific patterns that transfer poorly to diffusion, creating a notable generalization gap.

Taken together, these findings offer a comprehensive view of the current capabilities and limitations of both human and algorithmic deepfake detection systems. Humans consistently outperform automated models across a variety of metrics—including AUC, AP, d-prime, and precision—demonstrating not only superior accuracy but also greater consistency and robustness to variation in generative methods and video resolution. Moreover, the influence of prior AI experience highlights the role of digital literacy in enhancing detection performance, with experienced users showing both reduced error rates and stronger perceptual confidence in identifying manipulated media. While certain models like FFD \cite{dang2020detection} and UCF \cite{yan2023ucf} show promising results, the overall variability and occasional anomalies in model performance emphasize the need for further development of generalizable and interpretable detection algorithms. Ultimately, our results underscore the importance of integrating human insight and AI expertise in the broader fight against misinformation enabled by synthetic media.

\section{Conclusion}
Our study highlights a widening gap between the advancements in deepfake synthesis and the effectiveness of current detection methods. Despite progress in detection technologies, even the most robust models struggle to reliably identify increasingly sophisticated and realistic deepfakes. The emergence of advanced synthesis techniques, particularly those utilizing diffusion and NeRF methods, presents significant challenges that current detection frameworks are not fully equipped to handle. 

Moreover, while human evaluators demonstrate strong performance in identifying certain deepfakes, they too face difficulties with the most recent and high-quality forgeries. These findings emphasize the pressing need for further research and innovation in deepfake detection, as existing methods are rapidly becoming inadequate in the face of evolving synthetic media technologies. The continued advancement of detection strategies is crucial to maintaining the integrity of digital content and safeguarding against the growing threat posed by undetectable deepfakes.

\clearpage
\appendix
\addcontentsline{toc}{section}{Appendix}

\section{Extending Deepfake Detection Research through Comprehensive Performance Profiling}
To extend the scope and depth of our research on deepfake detection, we performed a comprehensive comparative analysis of ten state-of-the-art detection models across seven distinct deepfake generation methods and the results of the experiments are shown in Tables \ref{tab:comparison_table_full_auc_ap}, \ref{tab:comparison_table_full_precision_recall}, \ref{tab:comparison_table_full_dprime_C}. While our previous experiments were conducted on short 5-second video clips—common in existing literature—this study advances the field by evaluating model performance on full-length videos. This shift enables us to more accurately assess model behavior in realistic, temporally extended scenarios, where detection challenges such as motion consistency, long-term identity preservation, and temporal artifacts are more pronounced. Our evaluation spans a broad set of performance metrics, including AUC, average precision (AP), precision, recall, d-prime ($d'$), and decision criterion (C), thereby offering a rich, multidimensional perspective on model capabilities. Furthermore, by incorporating complex and diverse generation methods such as AniFaceDiff \cite{chen2024anifacediff}, FaceFusion \cite{facefusion2024}, and HyperReenact \cite{bounareli2023hyperreenact}, we provide insights into generalization performance under varied manipulation paradigms. This study not only benchmarks model robustness in practical conditions but also informs the development of next-generation forensic tools capable of operating reliably in real-world media ecosystems.

Among the metrics reported, AUC and AP provide a threshold-independent view of performance. Xception \cite{rossler2019faceforensics++}, CORE \cite{ni2022core}, and EfficientNet-B4 \cite{tan2019efficientnet} yield the highest AUCs across several generation methods, indicating reliable discrimination between real and fake samples. In contrast, models such as MesoNet \cite{afchar2018mesonet} and FFD \cite{dang2020detection} exhibit relatively lower AUCs and AP values, especially on more challenging manipulations like FaceFusion \cite{facefusion2024} and HyperReenact \cite{bounareli2023hyperreenact}, suggesting limited detection capability under complex generative distributions.

Precision and recall values further reveal model-specific trade-offs. Capsule \cite{nguyen2019capsule} and RECCE \cite{cao2022end} models show high precision but lower recall, reflecting a conservative detection strategy prone to fewer false positives but more false negatives. Conversely, models like UCF \cite{yan2023ucf} and SRM \cite{luo2021generalizing} prioritize higher recall, potentially at the cost of increased false positives. These trends are substantiated by F1 scores, where Xception \cite{rossler2019faceforensics++} and CORE \cite{ni2022core} again lead, suggesting a favorable balance between precision and recall.

From a signal detection theory perspective, $d'$ and C values offer insights into sensitivity and bias. Xception \cite{rossler2019faceforensics++} and EfficientNet-B4 \cite{tan2019efficientnet} consistently achieve high $d'$ values across most generators, indicating strong separability between classes. MesoNet \cite{afchar2018mesonet} and FFD \cite{dang2020detection} display both low $d'$ and erratic C values, indicating susceptibility to both sensitivity loss and decision bias. Models with near-zero C values, like CORE \cite{ni2022core} and EfficientNet-B4 \cite{tan2019efficientnet}, suggest an optimal trade-off between hit rate and false alarm rate.

Analyzing averages across generation methods reveals that certain models maintain stability, e.g., Xception \cite{rossler2019faceforensics++}, CORE \cite{ni2022core}, while others such as MesoNet \cite{afchar2018mesonet} and FFD \cite{dang2020detection} experience significant performance degradation, especially under newer generation techniques like AniFaceDiff \cite{chen2024anifacediff} and HyperReenact \cite{bounareli2023hyperreenact}. Standard deviation metrics further corroborate this, with high variance observed in MesoNet \cite{afchar2018mesonet} and SRM \cite{luo2021generalizing}, indicating model inconsistency across manipulation types. In contrast, the low standard deviation of Xception \cite{rossler2019faceforensics++} across metrics and methods underscores its adaptability.

This extensive evaluation suggests that models pretrained on large-scale, diverse datasets with deep architectures, e.g., Xception \cite{rossler2019faceforensics++}, EfficientNet-B4 \cite{tan2019efficientnet} generalize better to unseen or more complex manipulations. Simpler models, e.g., MesoNet \cite{afchar2018mesonet}, FFD) \cite{dang2020detection} may overfit to specific patterns or fail to capture subtle semantic inconsistencies introduced by advanced generation pipelines. These findings emphasize the necessity for robustness and diversity in training data, as well as the use of a broad set of performance metrics to accurately assess detection systems in real-world forensic deployments.

\begin{landscape}

\begin{table}[h!]
\centering
\def\arraystretch{1.2}
\caption{Comparison of Detection and Generation Performance (AUC and AP values) on Full Length Videos for Various Deepfake Detection Models (ROWS) and Generators (COLUMNS). The bolded values highlight the best overall detection model across rows and the most challenging deepfake generator across columns.}
\resizebox{1.35\textwidth}{!}{%
\begin{tabular}{lcccccccccccccccc}
\hline
\multirow{2}{*}{Model} &
  \multicolumn{2}{c}{AniFaceDiff \cite{chen2024anifacediff}} &
  \multicolumn{2}{c}{TPSMM \cite{zhao2022thin}} &
  \multicolumn{2}{c}{FaceVid \cite{wang2021one}} &
  \multicolumn{2}{c}{HyperReenact \cite{bounareli2023hyperreenact}} &
  \multicolumn{2}{c}{FaceFusion \cite{facefusion2024}} &
  \multicolumn{2}{c}{FADM \cite{zeng2023face}} &
  \multicolumn{2}{c}{SOTA \cite{xu2024vasa, ma2023dreamtalk, stypulkowski2024diffused, peng2024synctalk}} &
  \multicolumn{2}{c} {Detection Performance: Mean ± Std}\\ 
 & AUC & AP & AUC & AP & AUC & AP & AUC & AP & AUC & AP & AUC & AP & AUC & AP & AUC & AP \\ \hline
Capsule \cite{nguyen2019capsule} & 79.06 & 81.83 & 92.97 & 93.75 & 69.61 & 72.84 & 59.00 & 61.56 & 62.74 & 38.56 & 82.40 & 88.16 & 43.00 & 57.51 & 69.83 $\pm$ 16.66 & 70.6 $\pm$ 19.39 \\

CORE \cite{ni2022core} & 87.13 & 90.39 & 99.12 & 99.17 & 69.50 & 69.34 & 85.22 & 86.87 & 62.43 & 41.98 & 80.02 & 84.74 & 59.50 & 67.30 & 77.56 $\pm$ 14.38 & 77.11 $\pm$ 19.18 \\

EfficientNet-B4 \cite{tan2019efficientnet} & 85.96 & 88.34 & 91.37 & 92.96 & 64.12 & 66.80 & 39.46 & 47.52 & 43.81 & 34.68 & 70.43 & 79.71 & 48.25 & 67.56 & 63.34 $\pm$ 20.52 & 68.22 $\pm$ 21.23 \\

FFD \cite{dang2020detection} & 82.60 & 86.78 & 94.78 & 96.04 & 60.73 & 61.67 & 76.99 & 81.15 & 48.52 & 32.2 & 64.98 & 72.28 & 68.25 & 72.94 & 70.98 $\pm$ 15.22 & 71.87 $\pm$ 20.07 \\

MesoNet \cite{afchar2018mesonet} & 51.05 & 52.02 & 44.13 & 45.53 & 37.32 & 43.75 & 58.35 & 54.90 & 39.98 & 24.17 & 42.08 & 59.04 & 61.00 & 60.91 & 47.70 $\pm$ 9.25 & 48.62 $\pm$ 12.53 \\

Meso4Inception \cite{afchar2018mesonet} & 54.94 & 57.87 & 72.59 & 68.66 & 62.60 & 59.88 & 30.88 & 40.98 & 24.58 & 19.68 & 50.16 & 60.12 & 23.75 & 38.98 & 45.64 $\pm$ 19.41 & 49.45 $\pm$ 16.98 \\

RECCE \cite{cao2022end} & 88.72 & 91.81 & 96.22 & 97.01 & 63.93 & 68.08 & 77.90 & 83.47 & 64.87 & 51.05 & 72.96 & 82.01 & 51.75 & 63.55 & 73.76 $\pm$ 15.31 & 76.71 $\pm$ 16.43 \\

SRM \cite{luo2021generalizing} & 85.01 & 89.02 & 87.52 & 89.55 & 56.55 & 57.03 & 81.22 & 84.69 & 64.05 & 49.10 & 67.72 & 77.40 & 47.50 & 59.43 & 69.94 $\pm$ 15.20 & 72.32 $\pm$ 16.80 \\

UCF \cite{yan2023ucf} & 87.32 & 88.95 & 98.41 & 98.53 & 64.74 & 68.85 & 73.38 & 77.18 & 58.37 & 43.81 & 72.70 & 79.42 & 59.50 & 69.79 & 73.49 $\pm$ 14.8 & 75.22 $\pm$ 17.37 \\

Xception \cite{rossler2019faceforensics++} & 87.74 & 90.10 & 98.61 & 98.87 & 69.99 & 69.85 & 87.24 & 90.3 & 74.10 & 60.93 & 81.50 & 86.07 & 83.50 & 85.31 & \textbf{83.24} $\pm$ 9.44 & \textbf{83.06} $\pm$ 13.10 \\

Generation Performance: Mean ± Std & 78.95 $\pm$ 14.00 & 81.71 $\pm$ 14.43 & 87.57 $\pm$ 17.17 & 88.01 $\pm$ 17.44 & 61.91 $\pm$ 9.63 & 63.81 $\pm$ 8.63 & 66.96 $\pm$ 19.45 & 70.86 $\pm$ 17.98 & \textbf{54.35} $\pm$ 14.85 & \textbf{39.62} $\pm$ 12.54 & 68.50 $\pm$ 13.26 & 76.90 $\pm$ 10.19 & 54.60 $\pm$ 15.97 & 64.33 $\pm$ 11.97 \\ \hline 
\end{tabular}%
}
\label{tab:comparison_table_full_auc_ap}
\end{table}

\begin{table}[h!]
\centering
\def\arraystretch{1.2}
\caption{Comparison of Detection and Generation Performance (P, Precision and R, Recall Values) on Full Length Videos for Various Deepfake Detection Models (ROWS) and Generators (COLUMNS). The bolded values highlight the best overall detection model across rows and the most challenging deepfake generator across columns.}
\resizebox{1.35\textwidth}{!}{%
\begin{tabular}{lcccccccccccccccc}
\hline
\multirow{2}{*}{Model} &
  \multicolumn{2}{c}{AniFaceDiff \cite{chen2024anifacediff}} &
  \multicolumn{2}{c}{TPSMM \cite{zhao2022thin}} &
  \multicolumn{2}{c}{FaceVid \cite{wang2021one}} &
  \multicolumn{2}{c}{HyperReenact \cite{bounareli2023hyperreenact}} &
  \multicolumn{2}{c}{FaceFusion \cite{facefusion2024}} &
  \multicolumn{2}{c}{FADM \cite{zeng2023face}} &
  \multicolumn{2}{c}{SOTA \cite{xu2024vasa, ma2023dreamtalk, stypulkowski2024diffused, peng2024synctalk}} &
  \multicolumn{2}{c} {Detection Performance: Mean ± Std}\\ 
 & P & R & P & R & P & R & P & R & P & R & P & R & P & R & P & R \\ \hline

Capsule \cite{nguyen2019capsule} & 80.00 & 73.73 & 71.97 & 95.00 & 64.08 & 66.00 & 58.43 & 52.00 & 36.21 & 55.26 & 77.84 & 83.33 & 43.75 & 35.00 & 61.75 $\pm$ 16.78 & 65.76 $\pm$ 20.30\\

CORE \cite{ni2022core} & 67.00 & 77.01 & 93.27 & 97.00 & 81.58 & 31.00 & 89.55 & 60.00 & 56.25 & 23.68 & 91.25 & 46.79 & 87.50 & 35.00 & \textbf{80.91} $\pm$ 14.02 & 52.93 $\pm$ 26.67\\

EfficientNet-B4 \cite{tan2019efficientnet} & 69.00 & 76.67 & 88.17 & 82.00 & 76.09 & 35.00 & 50.00 & 11.00 & 42.11 & 21.05 & 83.58 & 35.90 & 100.00 & 35.00 & 72.71 $\pm$ 20.73 & 42.37 $\pm$ 26.88\\

FFD \cite{dang2020detection} & 80.00 & 74.07 & 72.73 & 96.00 & 59.09 & 52.00 & 67.27 & 74.00 & 23.40 & 28.95 & 72.09 & 59.62 & 75.00 & 45.00 & 64.23 $\pm$ 19.15 & 61.38 $\pm$ 22.10\\

MesoNet \cite{afchar2018mesonet} & 100.00 & 68.03 & 51.55 & 100.00 & 51.55 & 100.00 & 51.55 & 100.00 & 28.79 & 100.00 & 62.40 & 100.00 & 50.00 & 100.00 & 56.55 $\pm$ 21.64 & \textbf{95.43} $\pm$ 12.08\\

Meso4Inception \cite{afchar2018mesonet} & 85.00 & 62.96 & 53.80 & 99.00 & 52.51 & 94.00 & 40.14 & 57.00 & 20.56 & 57.89 & 63.36 & 94.23 & 45.71 & 80.00 & 51.58 $\pm$ 19.95 & 77.87 $\pm$ 18.42\\

RECCE \cite{cao2022end} & 80.00 & 81.22 & 85.19 & 92.00 & 70.37 & 38.00 & 79.75 & 63.00 & 52.94 & 47.37 & 81.82 & 46.15 & 83.33 & 25.00 & 76.20 $\pm$ 11.29 & 56.11 $\pm$ 23.94\\

SRM \cite{luo2021generalizing} & 86.00 & 75.44 & 68.18 & 90.00 & 53.85 & 49.00 & 66.67 & 84.00 & 40.00 & 73.68 & 71.62 & 67.95 & 53.85 & 35.00 & 62.88 $\pm$ 14.94 & 67.87 $\pm$ 19.48\\

UCF \cite{yan2023ucf} & 71.00 & 77.60 & 88.99 & 97.00 & 72.09 & 31.00 & 80.95 & 51.00 & 45.45 & 26.32 & 80.33 & 31.41 & 100.00 & 25.00 & 76.97 $\pm$ 17.10 & 48.48 $\pm$ 28.42\\

Xception \cite{rossler2019faceforensics++} & 76.00 & 76.38 & 81.15 & 99.00 & 69.33 & 52.00 & 77.67 & 80.00 & 47.73 & 55.26 & 81.60 & 65.38 & 84.62 & 55.00 & 74.01 $\pm$ 12.59 & 69.00 $\pm$ 17.14\\

Generation Performance: Mean ± Std & 79.40 $\pm$ 9.69 & 74.31 $\pm$ 5.22 & 75.50 $\pm$ 14.54 & 94.70 $\pm$ 5.46 & 65.05 $\pm$ 10.50 & 54.80 $\pm$ 24.86 & 66.20 $\pm$ 15.97 & 63.20 $\pm$ 24.11 & \textbf{39.34} $\pm$ 12.09 & 48.95 $\pm$ 25.19 & 76.59 $\pm$ 9.15 & 63.08 $\pm$ 23.75 & 72.38 $\pm$ 22.13 & \textbf{47.00} $\pm$ 24.74\\

 \hline 
\end{tabular}%
}
\label{tab:comparison_table_full_precision_recall}
\end{table}

\begin{table}[h!]
\centering
\def\arraystretch{1.2}
\caption{Comparison of Detection and Generation Performance ($d\text{'}$, Sensitivity Index and C, Decision Criterion Values) on Full Length Videos for Various Deepfake Detection Models (ROWS) and Generators (COLUMNS). The bolded values highlight the best overall detection model across rows and the most challenging deepfake generator across columns.}
\resizebox{1.35\textwidth}{!}{%
\begin{tabular}{lcccccccccccccccc}
\hline
\multirow{2}{*}{Model} &
  \multicolumn{2}{c}{AniFaceDiff \cite{chen2024anifacediff}} &
  \multicolumn{2}{c}{TPSMM \cite{zhao2022thin}} &
  \multicolumn{2}{c}{FaceVid \cite{wang2021one}} &
  \multicolumn{2}{c}{HyperReenact \cite{bounareli2023hyperreenact}} &
  \multicolumn{2}{c}{FaceFusion \cite{facefusion2024}} &
  \multicolumn{2}{c}{FADM \cite{zeng2023face}} &
  \multicolumn{2}{c}{SOTA \cite{xu2024vasa, ma2023dreamtalk, stypulkowski2024diffused, peng2024synctalk}} &
  \multicolumn{2}{c} {Detection Performance: Mean ± Std}\\ 
 & d' & C & d' & C & d' & C & d' & C & d' & C & d' & C & d' & C & d' & C \\ \hline

Capsule \cite{nguyen2019capsule} & 1.11 & -0.29 & 1.91 & -0.69 & 0.68 & -0.07 & 0.32 & 0.11 & 0.40 & 0.07 & 1.24 & -0.35 & -0.26 & 0.26 & 0.77 $\pm$ 0.71 & -0.14 $\pm$ 0.33\\

CORE \cite{ni2022core} & 1.88 & 0.50 & 3.32 & -0.22 & 0.95 & 0.97 & 1.70 & 0.59 & 0.73 & 1.08 & 1.36 & 0.76 & 1.26 & 1.02 & 1.60 $\pm$ 0.86 & 0.67 $\pm$ 0.45\\

EfficientNet-B4 \cite{tan2019efficientnet} & 1.69 & 0.35 & 2.11 & 0.14 & 0.80 & 0.79 & -0.04 & 1.21 & 0.39 & 1.00 & 0.83 & 0.78 & 5.98 & 3.37 & 1.68 $\pm$ 2.03 & \textbf{1.09} $\pm$ 1.07\\

FFD \cite{dang2020detection} & 1.14 & -0.27 & 2.05 & -0.73 & 0.35 & 0.12 & 0.94 & -0.17 & -0.26 & 0.43 & 0.54 & 0.03 & 0.91 & 0.58 & 0.81 $\pm$ 0.72 & 0.00 $\pm$ 0.44\\

MesoNet \cite{afchar2018mesonet} & 0.00 & -6.36 & 0.00 & -6.36 & 0.00 & -6.36 & 0.00 & -6.36 & 0.00 & -6.36 & 0.00 & -6.36 & 0.00 & -6.36 & 0.00 $\pm$ 0.00 & -6.36 $\pm$ 0.00\\

Meso4Inception \cite{afchar2018mesonet} & -0.27 & -1.17 & 1.02 & -1.82 & 0.25 & -1.43 & -1.13 & -0.74 & -1.11 & -0.75 & 0.27 & -1.44 & -0.80 & -1.24 & -0.25 $\pm$ 0.81 & -1.23 $\pm$ 0.39\\

RECCE \cite{cao2022end} & 1.75 & 0.04 & 2.36 & -0.23 & 0.65 & 0.63 & 1.29 & 0.31 & 0.89 & 0.51 & 0.86 & 0.52 & 0.97 & 1.16 & 1.25 $\pm$ 0.61 & 0.42 $\pm$ 0.45\\

SRM \cite{luo2021generalizing} & 1.21 & -0.47 & 1.42 & -0.57 & 0.11 & 0.08 & 1.13 & -0.43 & 0.77 & -0.25 & 0.6 & -0.17 & 0.14 & 0.45 & 0.77 $\pm$ 0.52 & -0.19 $\pm$ 0.36\\

UCF \cite{yan2023ucf} & 1.69 & 0.29 & 3.02 & -0.37 & 0.64 & 0.82 & 1.16 & 0.56 & 0.50 & 0.89 & 0.65 & 0.81 & 5.69 & 3.52 & \textbf{1.91} $\pm$ 1.89 & 0.93 $\pm$ 1.22\\

Xception \cite{rossler2019faceforensics++} & 1.40 & -0.01 & 3.02 & -0.82 & 0.74 & 0.32 & 1.53 & -0.08 & 0.82 & 0.28 & 1.09 & 0.15 & 1.41 & 0.58 & 1.43 $\pm$ 0.76 & 0.06 $\pm$ 0.45\\

Generation Performance: Mean ± Std & 1.16 $\pm$ 0.74 & -0.74 $\pm$ 2.03 & 2.02 $\pm$ 1.01 & \textbf{-1.17} $\pm$ 1.90 & 0.52 $\pm$ 0.32 & -0.41 $\pm$ 2.20 & 0.69 $\pm$ 0.88 & -0.50 $\pm$ 2.13 & \textbf{0.31} $\pm$ 0.62 & -0.31 $\pm$ 2.20 & 0.74 $\pm$ 0.42 & -0.53 $\pm$ 2.16 & 1.53 $\pm$ 2.38 & 0.33 $\pm$ 2.75\\ \hline 
\end{tabular}%
}
\label{tab:comparison_table_full_dprime_C}
\end{table}

\end{landscape}

\begin{landscape}

\section{Detailed Tabular Results}
The Tables \ref{tab:comparison_table_hr_auc_ap}, \ref{tab:comparison_table_lr_auc_ap} provide the complete quantitative results corresponding to the visual summaries presented in the main body of the paper. These tables report the performance of various deepfake detection models evaluated against multiple generation methods, with metrics including Area Under the Curve (AUC) and Average Precision (AP). All evaluations were conducted on 5-second video clips, with results stratified by resolution to assess robustness across different video qualities. Table \ref{tab:comparison_table_hr_auc_ap} presents the results on high-resolution videos, while Table \ref{tab:comparison_table_lr_auc_ap} reports performance on low-resolution videos. These detailed metrics support the main analysis and enable deeper examination of how detection performance varies across models, generation methods, and resolution levels.

\begin{table}[h!]
\centering
\def\arraystretch{1.2}
\caption{Comparison of Detection and Generation Performance (AUC and AP values) on 5-second High Resolution Videos for Various Deepfake Detection Models (ROWS) and Generators (COLUMNS). The bolded values highlight the best overall detection model across rows and the most challenging deepfake generator across columns.}
\resizebox{1.35\textwidth}{!}{%
\begin{tabular}{lcccccccccccccccc}
\hline
\multirow{2}{*}{Model} &
  \multicolumn{2}{c}{AniFaceDiff \cite{chen2024anifacediff}} &
  \multicolumn{2}{c}{TPSMM \cite{zhao2022thin}} &
  \multicolumn{2}{c}{FaceVid \cite{wang2021one}} &
  \multicolumn{2}{c}{HyperReenact \cite{bounareli2023hyperreenact}} &
  \multicolumn{2}{c}{FaceFusion \cite{facefusion2024}} &
  \multicolumn{2}{c}{FADM \cite{zeng2023face}} &
  \multicolumn{2}{c}{SOTA \cite{xu2024vasa, ma2023dreamtalk, stypulkowski2024diffused, peng2024synctalk}} &
  \multicolumn{2}{c} {Detection Performance: Mean ± Std}\\ 
 & AUC & AP & AUC & AP & AUC & AP & AUC & AP & AUC & AP & AUC & AP & AUC & AP & AUC & AP \\ \hline
Human & 99.75 & 99.76 & 93.50 & 95.67 & 91.58 & 93.93 & 98.00 & 98.45 & 95.75 & 96.63 & 99.25 & 99.25 & 97.00 & 96.87 & \textbf{96.40} $\pm$ 3.01 & \textbf{97.22} $\pm$ 2.07 \\
Capsule \cite{nguyen2019capsule} & 47.5 & 48.59 & 74.00 & 75.67 & 80.00 & 78.19 & 65.50 & 62.47 & 66.00 & 65.91 & 85.75 & 81.38 & 57.75 & 60.24 & 68.07 $\pm$ 13.11 & 67.49 $\pm$ 11.63 \\
CORE \cite{ni2022core} & 46.25 & 49.33 & 84.50 & 82.03 & 70.50 & 67.00 & 84.25 & 81.22 & 69.50 & 71.72 & 83.00 & 79.72 & 65.50 & 70.02 & 71.93 $\pm$ 13.79 & 71.58 $\pm$ 11.45 \\
EfficientNet-B4 \cite{tan2019efficientnet} & 45.50 & 46.88 & 65.50 & 71.09 & 69.25 & 70.66 & 69.00 & 66.52 & 67.00 & 73.85 & 77.25 & 74.41 & 60.00 & 70.82 & 64.79 $\pm$ 9.94 & 67.75 $\pm$ 9.55 \\
FFD \cite{dang2020detection} & 50.25 & 50.40 & 70.00 & 70.66 & 63.75 & 66.59 & 67.75 & 66.24 & 81.50 & 82.26 & 79.00 & 81.48 & 76.50 & 74.31 & 69.82 $\pm$ 10.72 & 70.28 $\pm$ 10.88 \\
MesoNet \cite{afchar2018mesonet} & 41.50 & 49.05 & 35.25 & 41.04 & 46.00 & 50.59 & 85.50 & 86.78 & 46.00 & 51.20 & 41.25 & 50.85 & 67.75 & 73.62 & 51.89 $\pm$ 18.02 & 57.59 $\pm$ 16.28 \\
Meso4Inception \cite{afchar2018mesonet} & 53.50 & 54.54 & 67.75 & 64.51 & 72.00 & 71.50 & 57.50 & 53.54 & 67.25 & 64.56 & 80.75 & 74.19 & 71.75 & 68.2 & 67.21 $\pm$ 9.21 & 64.43 $\pm$ 7.91 \\
RECCE \cite{cao2022end} & 39.75 & 45.38 & 81.75 & 84.56 & 68.00 & 71.00 & 83.75 & 84.47 & 77.00 & 84.98 & 85.50 & 86.81 & 62.50 & 70.34 & 71.18 $\pm$ 16.24 & 75.36 $\pm$ 14.91 \\
SRM \cite{luo2021generalizing} & 36.75 & 43.62 & 73.00 & 68.57 & 61.50 & 59.62 & 67.50 & 61.24 & 74.00 & 78.12 & 77.50 & 73.73 & 57.00 & 62.59 & 63.89 $\pm$ 13.98 & 63.93 $\pm$ 11.25 \\
UCF \cite{yan2023ucf} & 53.50 & 53.77 & 81.50 & 82.36 & 68.750 & 72.70 & 76.75 & 74.08 & 79.00 & 81.91 & 83.00 & 80.33 & 75.50 & 77.30 & 74.00 $\pm$ 10.16 & 74.64 $\pm$ 9.93 \\
Xception \cite{rossler2019faceforensics++} & 41.00 & 44.10 & 75.00 & 74.09 & 71.25 & 65.01 & 82.00 & 74.16 & 77.50 & 81.34 & 85.50 & 78.83 & 78.00 & 72.30 & 72.89 $\pm$ 14.8 & 69.98 $\pm$ 12.53 \\
Generation Performance: Mean ± Std & \textbf{50.48} $\pm$ 17.24 & \textbf{53.22} $\pm$ 15.83 & 72.89 $\pm$ 14.91 & 73.66 $\pm$ 13.94 & 69.33 $\pm$ 11.22 & 69.71 $\pm$ 10.86 & 76.14 $\pm$ 11.75 & 73.56 $\pm$ 13.23 & 72.77 $\pm$ 12.34 & 75.68 $\pm$ 12.18 & 79.80 $\pm$ 14.14 & 78.27 $\pm$ 11.59 & 69.93 $\pm$ 11.70 & 72.42 $\pm$ 9.50 \\ \hline 
\end{tabular}%
}
\label{tab:comparison_table_hr_auc_ap}
\end{table}

\begin{table}[h!]
\centering
\def\arraystretch{1.2}
\caption{Comparison of Detection and Generation Performance (AUC and AP values) on 5-second Low Resolution Videos for Various Deepfake Detection Models (ROWS) and Generators (COLUMNS). The bolded values highlight the best overall detection model across rows and the most challenging deepfake generator across columns.}
\resizebox{1.35\textwidth}{!}{%
\begin{tabular}{lcccccccccccccccc}
\hline
\multirow{2}{*}{Model} &
  \multicolumn{2}{c}{AniFaceDiff \cite{chen2024anifacediff}} &
  \multicolumn{2}{c}{TPSMM \cite{zhao2022thin}} &
  \multicolumn{2}{c}{FaceVid \cite{wang2021one}} &
  \multicolumn{2}{c}{HyperReenact \cite{bounareli2023hyperreenact}} &
  \multicolumn{2}{c}{FaceFusion \cite{facefusion2024}} &
  \multicolumn{2}{c}{FADM \cite{zeng2023face}} &
  \multicolumn{2}{c}{SOTA \cite{xu2024vasa, ma2023dreamtalk, stypulkowski2024diffused, peng2024synctalk}} &
  \multicolumn{2}{c} {Detection Performance: Mean ± Std}\\ 
 & AUC & AP & AUC & AP & AUC & AP & AUC & AP & AUC & AP & AUC & AP & AUC & AP & AUC & AP \\ \hline
Human & 98.50 & 98.75 & 87.50 & 92.00 & 84.00 & 87.80 & 99.75 & 99.76 & 88.50 & 92.09 & 76.00 & 83.14 & 83.50 & 87.10 & \textbf{88.25} $\pm$ 7.83 & \textbf{91.52} $\pm$ 5.66 \\
Capsule \cite{nguyen2019capsule} & 53.25 & 53.83 & 70.75 & 73.91 & 84.25 & 82.80 & 81.00 & 74.77 & 67.50 & 62.72 & 89.50 & 87.51 & 37.00 & 45.87 & 69.04 $\pm$ 18.59 & 68.77 $\pm$ 15.25 \\
CORE \cite{ni2022core} & 55.75 & 60.86 & 73.25 & 75.57 & 79.00 & 79.86 & 87.00 & 88.94 & 78.50 & 79.81 & 87.75 & 86.94 & 39.25 & 54.02 & 71.50 $\pm$ 17.80 & 75.14 $\pm$ 13.06 \\
EfficientNet-B4 \cite{tan2019efficientnet} & 58.75 & 63.26 & 63.50 & 72.50 & 78.75 & 82.00 & 75.75 & 78.22 & 69.25 & 75.25 & 87.50 & 89.45 & 42.75 & 62.52 & 68.04 $\pm$ 14.72 & 74.74 $\pm$ 9.73 \\
FFD \cite{dang2020detection} & 58.75 & 61.58 & 62.25 & 65.99 & 72.00 & 77.33 & 75.25 & 78.70 & 77.25 & 78.77 & 81.50 & 82.79 & 65.50 & 67.42 & 70.36 $\pm$ 8.39 & 73.23 $\pm$ 8.07 \\
MesoNet \cite{afchar2018mesonet} & 41.50 & 50.08 & 29.25 & 38.74 & 39.25 & 44.48 & 81.50 & 83.64 & 47.75 & 55.77 & 36.50 & 46.88 & 57.50 & 61.65 & 47.61 $\pm$ 17.39 & 54.46 $\pm$ 14.88 \\
Meso4Inception \cite{afchar2018mesonet} & 54.25 & 55.22 & 57.50 & 60.17 & 75.25 & 73.81 & 66.25 & 62.66 & 65.75 & 63.69 & 77.25 & 69.69 & 39.75 & 49.38 & 62.29 $\pm$ 13.01 & 62.09 $\pm$ 8.27 \\
RECCE \cite{cao2022end} & 52.00 & 61.43 & 76.25 & 83.68 & 77.75 & 82.03 & 95.75 & 96.32 & 83.25 & 86.28 & 86.75 & 89.23 & 40.25 & 58.94 & 73.14 $\pm$ 19.83 & 79.70 $\pm$ 14.12 \\
SRM \cite{luo2021generalizing} & 49.75 & 52.58 & 61.25 & 63.63 & 69.00 & 71.15 & 77.50 & 75.97 & 72.50 & 75.59 & 77.75 & 78.33 & 41.00 & 56.35 & 64.11 $\pm$ 14.19 & 67.66 $\pm$ 10.24 \\
UCF \cite{yan2023ucf} & 60.00 & 62.64 & 68.50 & 69.71 & 75.75 & 78.43 & 82.25 & 79.32 & 77.50 & 75.45 & 81.50 & 76.93 & 60.50 & 67.10 & 72.29 $\pm$ 9.38 & 72.80 $\pm$ 6.37 \\
Xception \cite{rossler2019faceforensics++} & 45.75 & 49.50 & 56.00 & 61.61 & 70.50 & 72.57 & 69.75 & 68.14 & 72.75 & 76.01 & 78.25 & 78.64 & 59.00 & 62.29 & 64.57 $\pm$ 11.36 & 66.97 $\pm$ 10.05 \\
Average - Generation Model & 57.11 $\pm$ 14.85 & \textbf{60.88} $\pm$ 13.57 & 64.18 $\pm$ 14.76 & 68.86 $\pm$ 13.82 & 73.23 $\pm$ 12.30 & 75.66 $\pm$ 11.46 & 81.07 $\pm$ 10.12 & 80.59 $\pm$ 11.12 & 72.77 $\pm$ 10.71 & 74.68 $\pm$ 10.51 & 78.20 $\pm$ 14.65 & 79.05 $\pm$ 12.26 & \textbf{51.45} $\pm$ 14.85 & 61.15 $\pm$ 10.97 \\ \hline 
\end{tabular}%
}
\label{tab:comparison_table_lr_auc_ap}
\end{table}
    
\end{landscape}

\section{Detection Performance Across Video Conditions}
\begin{figure*}
    \centering
    \includegraphics[width=0.92\textwidth]{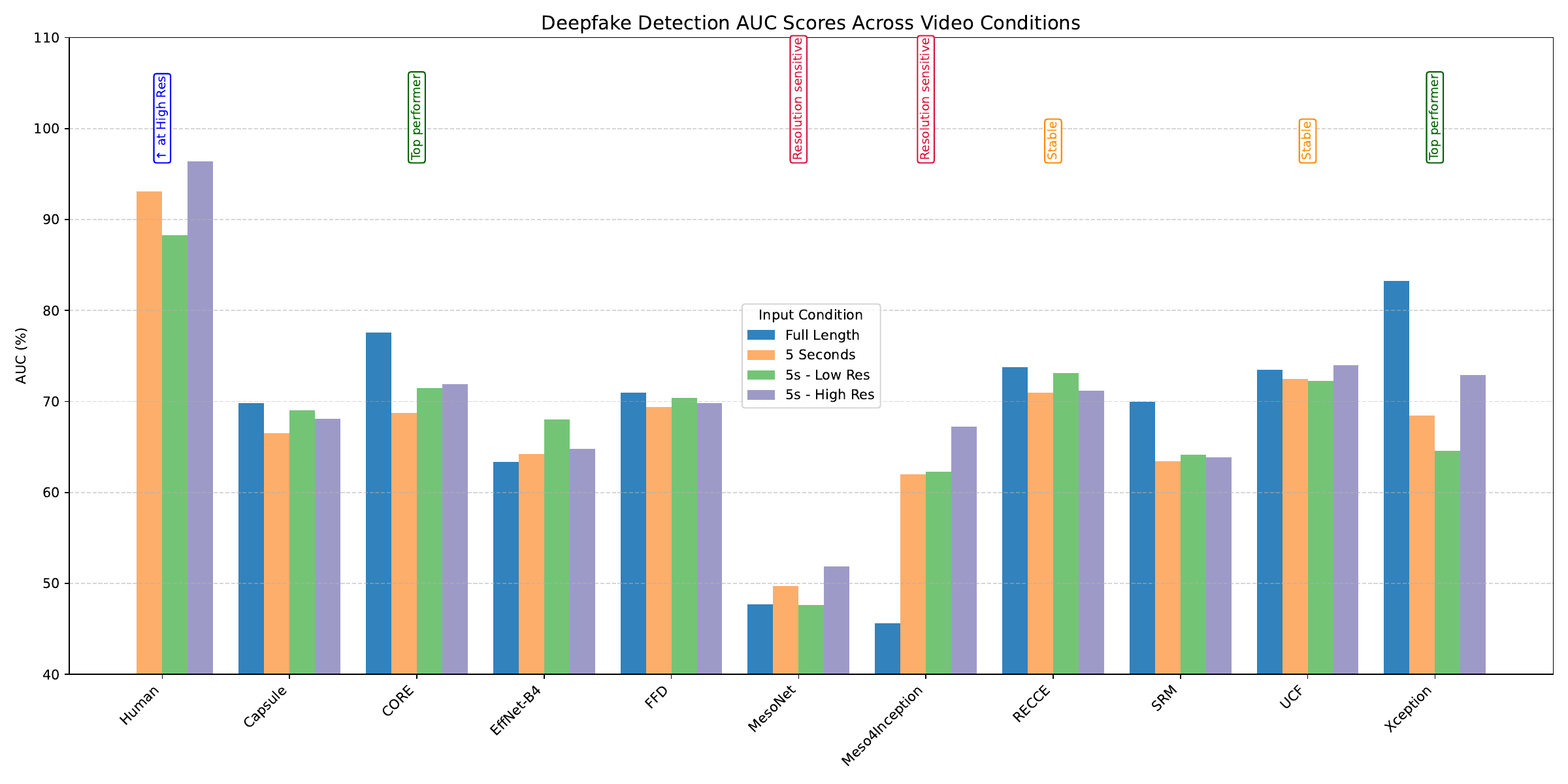}
    \caption{Average AUC (\%) scores of deepfake detection models across different video input conditions: full-length, 5-second clips, and 5-second clips at low and high resolutions. Xception \cite{rossler2019faceforensics++} and CORE \cite{ni2022core} exhibit top performance overall, with Xception \cite{rossler2019faceforensics++} achieving the highest AUC on full-length videos. Human detection accuracy increases with resolution, a trend not mirrored by most automated models. MesoNet \cite{afchar2018mesonet} and Meso4Inception \cite{afchar2018mesonet} show high sensitivity to resolution changes, while RECCE \cite{cao2022end} and UCF \cite{yan2023ucf} maintain stable performance across all conditions.}
    \label{fig:auc_scores_comparison_varying_condition}
\end{figure*}

Table~\ref{tab:detection-results} and Figure~\ref{fig:auc_scores_comparison_varying_condition} present the average AUC (\%) scores of deepfake detection models evaluated under varying input conditions, specifically comparing performance on full-length videos, 5-second clips, and 5-second clips at both low and high resolutions. These results reflect the mean detection performance of each model across multiple generative methods. Notably, human evaluators were only assessed on short clips and not on full-length videos due to practical limitations. Among automated models, Xception \cite{rossler2019faceforensics++} and CORE \cite{ni2022core} demonstrate the strongest overall performance, with Xception achieving the highest AUC on full-length inputs. This performance gain on full videos can be attributed to the richer temporal context and greater opportunity to observe forgery artifacts across frames, which enables models to extract more robust and representative features. Human detection accuracy improves noticeably with resolution, indicating that visual clarity significantly aids human judgment. In contrast, this resolution-dependent gain is not consistently reflected in automated models. For instance, MesoNet \cite{afchar2018mesonet} and Meso4Inception \cite{afchar2018mesonet} show substantial performance drops under low-resolution conditions, highlighting their sensitivity to input quality. On the other hand, RECCE \cite{cao2022end} and UCF \cite{yan2023ucf} display relatively stable AUC scores across all conditions, indicating robustness to both duration and resolution variations. These findings underscore the need for developing detection models that maintain high performance across a broad range of real-world input scenarios. In practice, such scenarios often involve substantial variations in data quality, environmental conditions, and content characteristics that are not fully represented in controlled benchmark datasets. Consequently, robust models should be designed to generalize effectively under diverse and potentially unseen conditions, ensuring reliable performance when deployed in real-world applications.

\begin{table*}[ht]
\centering
\caption{Average AUC (\%) of Deepfake Detection Models Across Varying Video Durations and Resolutions. 
Each value represents the model's mean performance across multiple generative methods. 
Detection performance is evaluated on full-length videos, 5-second clips, and 5-second clips at low and high resolutions. The bolded values highlight the best detection model on each video condition. \\\textit{Note: Human evaluation was not conducted on full-length videos; corresponding value is not available.}}

\label{tab:detection-results}
\begin{tabular}{lcccc}
\toprule
\textbf{Detection Model} & \textbf{Full Length} & \textbf{5 Seconds} & \textbf{5s - Low Res} & \textbf{5s - High Res} \\
\midrule
Human              & --    & \textbf{93.10} & \textbf{88.25} & \textbf{96.40} \\
Capsule \cite{nguyen2019capsule}            & 69.83 & 66.55 & 69.04 & 68.07 \\
CORE \cite{ni2022core}               & 77.56 & 68.74 & 71.50 & 71.93 \\
EfficientNet-B4 \cite{tan2019efficientnet}    & 63.34 & 64.24 & 68.04 & 64.79 \\
FFD \cite{dang2020detection}               & 70.98 & 69.38 & 70.36 & 69.82 \\
MesoNet \cite{afchar2018mesonet}           & 47.70 & 49.73 & 47.61 & 51.89 \\
Meso4Inception \cite{afchar2018mesonet}     & 45.64 & 61.98 & 62.29 & 67.21 \\
RECCE \cite{cao2022end}           & 73.76 & 70.94 & 73.14 & 71.18 \\
SRM \cite{luo2021generalizing}               & 69.94 & 63.43 & 64.11 & 63.89 \\
UCF  \cite{yan2023ucf}              & 73.49 & 72.49 & 72.29 & 74.00 \\
Xception  \cite{rossler2019faceforensics++}         & \textbf{83.24} & 68.43 & 64.57 & 72.89 \\
\bottomrule
\end{tabular}
\end{table*}

\section{Detection Hardness and Generalizability: A Theoretical Perspective}
Before delving into the distributional analysis, it is worth contrasting the generative objectives that underlie GAN- and diffusion-based deepfake synthesis, as these influence the statistical properties of their outputs and thus detection difficulty. GAN-based models optimize a minimax objective
\[
\min_G \max_D \ \mathbb{E}_{x \sim P_r}[\log D(x)] + \mathbb{E}_{z \sim p_z}[\log(1 - D(G(z)))],
\]
where the generator $G$ maps random noise $z$ to a sample in a single pass, and the discriminator $D$ learns to distinguish real from fake. This adversarial process can yield realistic outputs but is prone to mode collapse and generator-specific artifacts. In contrast, diffusion models minimize a denoising score-matching objective:
\[
\mathcal{L}_{\mathrm{diff}} = \mathbb{E}_{x_0, \epsilon, t} \left[ \big\| \epsilon - \epsilon_\theta(x_t, t) \big\|_2^2 \right],
\]
where the network predicts the noise added to a gradually corrupted sample. This iterative refinement better matches the real data distribution and produces fewer structured artifacts. In practice, the GAN loss can lead to visually sharp but sometimes unstable generations with detectable high-frequency patterns in the frequency domain, whereas the diffusion loss yields smoother and more statistically consistent samples that lack such periodic artifacts, making them less conspicuous to frequency-based detection methods. These differences in objectives help explain the smaller feature-space distances and higher Bayes errors observed for diffusion-generated content, as analyzed below.

We assess distributional closeness between real and generated content in a fixed feature space using Fréchet Inception Distance (FID) and Fréchet Video Distance (FVD) \cite{chen2024anifacediff}. 
For HyperReenact (GAN) and AniFaceDiff (diffusion), we obtain $\sqrt{\mathrm{FID}}$: $9.95$ vs.\ $6.02$ (ratio $0.605$) and $\sqrt{\mathrm{FVD}}$: $19.10$ vs.\ $15.55$ (ratio $0.814$), indicating that diffusion samples are $39.5\%$ (images) and $18.6\%$ (videos) closer to real than GAN samples on the Wasserstein scale \cite{panaretos2019statistical}. 
Under standard $T_2$-type concentration inequalities in the embedding space, the total variation distance satisfies $\mathrm{TV}(P_r,P_f) \le \sqrt{\mathrm{FID}(P_r,P_f)}/(2\sigma_\phi)$ (and analogously for FVD), where $\sigma_\phi$ characterizes feature concentration. 
Since $\mathrm{FID}_{\text{diffusion}} < \mathrm{FID}_{\text{GAN}}$ and $\mathrm{FVD}_{\text{diffusion}} < \mathrm{FVD}_{\text{GAN}}$, it follows that $\mathrm{TV}(P_r,P_d) < \mathrm{TV}(P_r,P_g)$, which by the Bayes error formula 
\[
\varepsilon^*(P_r,P_f) = \frac{1-\mathrm{TV}(P_r,P_f)}{2}
\]
implies $\varepsilon^*(P_r,P_d) > \varepsilon^*(P_r,P_g)$. 
Therefore, even the optimal detector will make more errors on diffusion-generated content, confirming that AniFaceDiff is intrinsically harder to detect than HyperReenact.

When a detector is trained to discriminate $P_r$ from $P_g$, it often learns generator-specific artifacts (e.g., high-frequency spikes in GAN images) that are not present in $P_d$. 
From a domain adaptation perspective, the target risk on real vs.\ diffusion satisfies
\[
R_{r\cup d}(h) \ \le\ R_{r\cup g}(h) \ + \ \frac{1}{2} \, d_{\mathcal{H}\Delta\mathcal{H}}(P_g,P_d) \ + \ \lambda,
\]
where $d_{\mathcal{H}\Delta\mathcal{H}}$ is the discrepancy distance between GAN and diffusion fakes in the learned feature space and $\lambda$ is the joint optimal risk. 
Because $P_d$ lies statistically closer to $P_r$ than $P_g$ does, features that separate $P_r$ from $P_g$ are less transferable to $P_d$, increasing the discrepancy term and degrading performance. 
This mathematical relationship explains the empirical observation that detectors trained solely on GAN-based fakes generalize poorly to diffusion-based fakes.

{\small
\bibliographystyle{IEEEtran}
\bibliography{references}
}
\end{document}